%% file: main.tex
\documentclass{nature}

\usepackage[sort&compress,super]{natbib}

\usepackage[font=small]{caption}
\usepackage{xcolor}
\usepackage{media9}
\usepackage{comment}
\usepackage{graphicx}
\usepackage{algorithm}
\usepackage{algorithmic}
\usepackage[utf8]{inputenc} % allow utf-8 input
\usepackage[T1]{fontenc}    % use 8-bit T1 fonts
\usepackage{hyperref}       % hyperlinks
\usepackage{url}            % simple URL typesetting
\usepackage{booktabs}       % professional-quality tables
\usepackage{amsfonts}       % blackboard math symbols
\usepackage{nicefrac}       % compact symbols for 1/2, etc.
\usepackage{microtype}      % microtypography
\usepackage{float}
\usepackage{amsmath,amsfonts, amssymb, makecell}
\usepackage{mathtools}
\newcommand{\defeq}{\vcentcolon=}

\usepackage{multirow}
\usepackage{textcomp}

\definecolor{revisedcolor}{RGB}{0, 0, 0}
\newcommand{\floor}[1]{\left\lfloor #1 \right\rfloor}

\newcommand{\john}[1]{{{\textcolor{black}{#1}}}}
\newcommand{\Carla}[1]{{{\textcolor{black}{#1}}}}
\newcommand{\Carlab}[1]{{{\textcolor{black}{#1}}}}
\newcommand{\Carlad}[1]{{{\textcolor{black}{#1}}}}
\newcommand{\Carlanew}[1]{{{\textcolor{black}{#1}}}}
\newcommand{\revised}[1]{{{\textcolor{black}{#1}}}}
\DeclareCaptionStyle{mystyle}%
[margin=5mm,justification=centering]% 
{font=footnotesize,labelfont=sc,margin={10mm,0mm}}
\usepackage{hyperref}
\hypersetup{
    colorlinks=true,
    linkcolor=blue,
    filecolor=blue,      
    urlcolor=cyan,
    citecolor=blue,
}

\usepackage{multibib}

\usepackage{etoolbox}
\makeatletter
\patchcmd{\thebibliography}{%
  \section*{\refname}\@mkboth{\MakeUppercase\refname}{\MakeUppercase\refname}%
}{}{}{}
\makeatother

\newcites{M}{ }
\newcites{S}{ } 

\makeatletter
\let\saved@includegraphics\includegraphics
\AtBeginDocument{\let\includegraphics\saved@includegraphics}
\renewenvironment*{figure}{\@float{figure}}{\end@float}
\makeatother

\title{\small\Carlanew{Automating Crystal-Structure Phase Mapping: Combining Deep Learning with Constraint Reasoning}}

\author{Di Chen$^{1},$  Yiwei Bai$^{1},$ Sebastian Ament$^{1},$ Wenting Zhao$^{1},$ Dan Guevarra$^{2},$ Lan Zhou$^{2},$  Bart Selman$^{1},$ R.\ Bruce van Dover$^{3},$ John M. Gregoire$^{2,4,\dag}$\& Carla P.\ Gomes$^{1,\dag}$}

\begin{document}
\maketitle
\begin{affiliations}
 \item [\footnotesize{$1$}]\ Cornell University, Department of Computer Science
\item [\footnotesize{$2$}]\  California Institute of Technology, Joint Center for Artificial Photosynthesis
 \item [\footnotesize{$3$}]\ Cornell University, Department of Materials  Science and Engineering
 \item [\footnotesize{$4$}]\  California Institute of Technology, Division of Engineering and Applied Science
\item [\footnotesize{$\dag$}]\ Corresponding authors: gomes@cs.cornell.edu and gregoire@caltech.edu.
 %\item Corresponding authors: gomes@cs.cornell.edu, gregoire@caltech.edu, or  daniel.fink@cornell.edu.
\end{affiliations}

%\begin{affiliations}
% \item Cornell University, Department of Computer Science
%\item California Institute of Technology
 %\item Corresponding authors: gomes@cs.cornell.edu and  gregoire@caltech.edu.
%\end{affiliations}

%uncomment then next commented out text
%\thispagestyle{empty}
\input{abstract}
%uncomment then next commented out text
%\thispagestyle{empty}
\input{main-intro}

\input{drnets}

\input{main-sudoku}

\input{main-phase-mapping}

\input{new-discussion}
\input{methods}

%\newbibstartnumber{31}

%methods go before figure but I'm changing that for now
%\input{main-figures}
%we inlined the figures
%moved bib here 
\newpage
\begin{addendum}
 \item [Acknowledgements]The development of DRNets %DRNETs
 was supported by NSF awards CCF-1522054 (Expeditions in computing), and CNS-1059284 (Infrastructure). DRNets for phase mapping and corresponding experimental work were also supported by AFOSR Multidisciplinary University Research Initiatives (MURI) Program FA9550-18-1-0136, ARO awards W911NF-14-1-0498 and W911NF-17-1-0187,
 %US DOE Award No. DE-SC0020383, 
 and an award from the Toyota Research Institute. Solar fuels experiments were supported by US DOE Award No. DE-SC0004993 and solar photochemistry analysis in the context of the DRNets solution was supported by US DOE Award No. DE-SC0020383. Use of SSRL is supported by DOE Contract No. DE-AC02-76SF00515. 
 The authors also thank Junwen Bai for assistance with running the IAFD baseline, Aniketa Shinde for photoelectrochemistry experiments, and Rich Berstein for assistance with figure generation.
 %, and Chris Wood and Ian Davies for reviewing the DMVP-DRNets eBird results. 
 %We also thank the anonymous referees for valuable feedback, which helped us  significantly improve the manuscript.
 \item[Competing Interests] The authors declare that they have no competing financial interests.
  \item[Data availability] MNIST-Sudoku and  crystal-structure mapping data 
will be avaialble for  downloaded from: \Carlanew{Data provided in zip file  with supplementary files. Please do not distribute it (confidential). GitHub link to be provided if/when manuscript accepted.} %https://www.cs.cornell.edu/gomes/udiscoverit/. 
%  Non-sensitive eBird data are freely available on the eBird website https://ebird.org/science/download-ebird-data-products. This study used checklists collected within North America from Jan 1 2004 – Dec 31 2018. Further details of checklists used for the analysis are in the  \hyperref[sec:si]{Supplementary
  \item[Contributions] C.P.G.\ conceived and managed the overall study. J.M.G.\ and C.P.G.\ conceived and managed the crystal-structure phase mapping project.
  %D.F.\ and C.P.G.\ conceived and managed the bird-species project. 
  D.C.\ and C.P.G.\  conceived the MNIST-SUDOKU project.   D.C.\ and C.P.G.\ conceptualized Deep Reasoning Nets. D.C.\ developed and implemented DRNets,  in particular, DRNets for MVP, MNIST-Sudoku, and Crystal-Structure Phase mapping. 
  %D.C.\ implemented and ran the JSDM baselines. Y.B. ran DRNets with the eBird data. 
  Y.B.\ performed the large-scale experiments, assisted on implementing DRNets for MNIST-Sudoku, and baseline comparisons for MNIST-Sudoku. S.A.\ performed background subtraction for the Bi-Cu-V-O system. W.Z.\ implemented baselines for crystal-structure phase mapping and assisted on generating MNIST-Sudoku data. 
  %D.F., O.R., and V.R.G.\ processed JSDM data, made substantial contributions to the analysis and interpretation of results. 
  L.Z. and D.G. generated phase mapping datasets and interpreted/validated solutions. C.P.G., D. C., and J.M.G.\ were the main authors of the manuscript with contributions from B.S.\ and R.B.v.D.\ and comments from all authors.
   \item[Code availability] All relevant code  will be   available from: 
   \Carlanew{Code provided in zip file with supplementary file. Please do not distribute it (confidential). GitHub link to be provided with final version of manuscript. } 
   %https://www.cs.cornell.edu/gomes/udiscoverit/.
 \item[Correspondence] Correspondence and requests for materials should be addressed to:\\ C.P.G.~(gomes@cs.cornell.edu) and  J.M.G.~(gregoire@caltech.edu).
\end{addendum}
\section*{\small{References}}
\bibliographystyle{unsrt}
\bibliography{nature}
\newpage
%\bibliography{nature}
%\addtocounter{\@listctr}{30}
% {\removebibheader
%  \bibliographynovels{M}
% }
%\bibliographystyleM{unsrt}
%\bibliographyM{nature}
% \bibliographyM{nature}
\input{extended-data}

\input{si}
% \let\oldthebibliography=\thebibliography
% \renewenvironment{thebibliography}[1]{%
%   \oldthebibliography{#1}%
%   \setcounter{enumiv}{0}%
% }
%\newbibstartnumber{1}

\section*{\small{References in Supplementary Methods}}
\bibliographystyleS{unsrt}
\bibliographyS{nature}
\end{document}

%% file: abstract.tex
\begin{abstract}
{\small Crystal-structure phase mapping is a core, long-standing challenge in materials science that requires identifying crystal structures, or mixtures thereof, in synthesized materials. Materials science experts excel at solving  simple systems but cannot solve complex systems, creating a major bottleneck in high-throughput materials discovery. Herein we show how to automate crystal-structure phase mapping. We formulate phase mapping as an unsupervised pattern demixing problem and describe how to solve it using  Deep Reasoning Networks (DRNets). DRNets combine deep learning with constraint reasoning for incorporating scientific prior knowledge and consequently require only a modest amount of (unlabeled) data. DRNets compensate for the limited data by exploiting and magnifying the rich prior-knowledge about the thermodynamic rules governing the mixtures of crystals with constraint reasoning seamlessly integrated into neural network optimization. DRNets are designed with an interpretable latent space for encoding prior-knowledge domain constraints and seamlessly integrate constraint reasoning into neural network optimization. DRNets surpass previous approaches on crystal-structure phase mapping, unraveling the Bi-Cu-V oxide phase diagram, and aiding the discovery of solar-fuels materials. 
}

\end{abstract}

%% file: main-intro.tex
Artificial Intelligence (AI)\cite{stajic_rise_2015}
aims to develop intelligent systems, inspired in part by human intelligence. AI systems are now performing at human and even superhuman levels on a range of tasks such as image identification \cite{szegedy2015going}, 
face,~\cite{taigman2014deepface} and speech recognition \cite{graves2013speech}.
AI also has  the potential to 
accelerate scientific discovery  dramatically.\cite{
%AI-Science-2015,
gil2014amplify,ball2018learning,tabor2018accelerating,sanchez2018inverse,sun_accelerated_2019,kusne_--fly_2020}
Recent AI achievements have been driven mainly by advances in  supervised
deep %\Carla{supervised}
learning\cite{lecun2015deep}, 
which requires  large labeled datasets to supervise model training. However, in general,  scientists do not have large amounts of labeled data for scientific discovery.
 They often solve complex tasks using only a few data samples by amplifying intuitive pattern recognition with detailed reasoning about prior knowledge to make sense of the data. Such a  hybrid strategy has been difficult to automate.
 \Carlanew{Herein we consider crystal-structure phase mapping, a long standing challenge in materials science
 that is emblematic of the class of scientific problems whose automation constitutes a substantial advancement with respect %to this grand challenge.
 to the grand challenge of high-throughput unsupervised  scientific data interpretation.}
 
%However, scientific discovery often requires combining data analysis with reasoning  
%about prior knowledge, which is still a challenge for AI.
\begin{figure}[th]
\centering
  \includegraphics [width=0.9
  \linewidth]{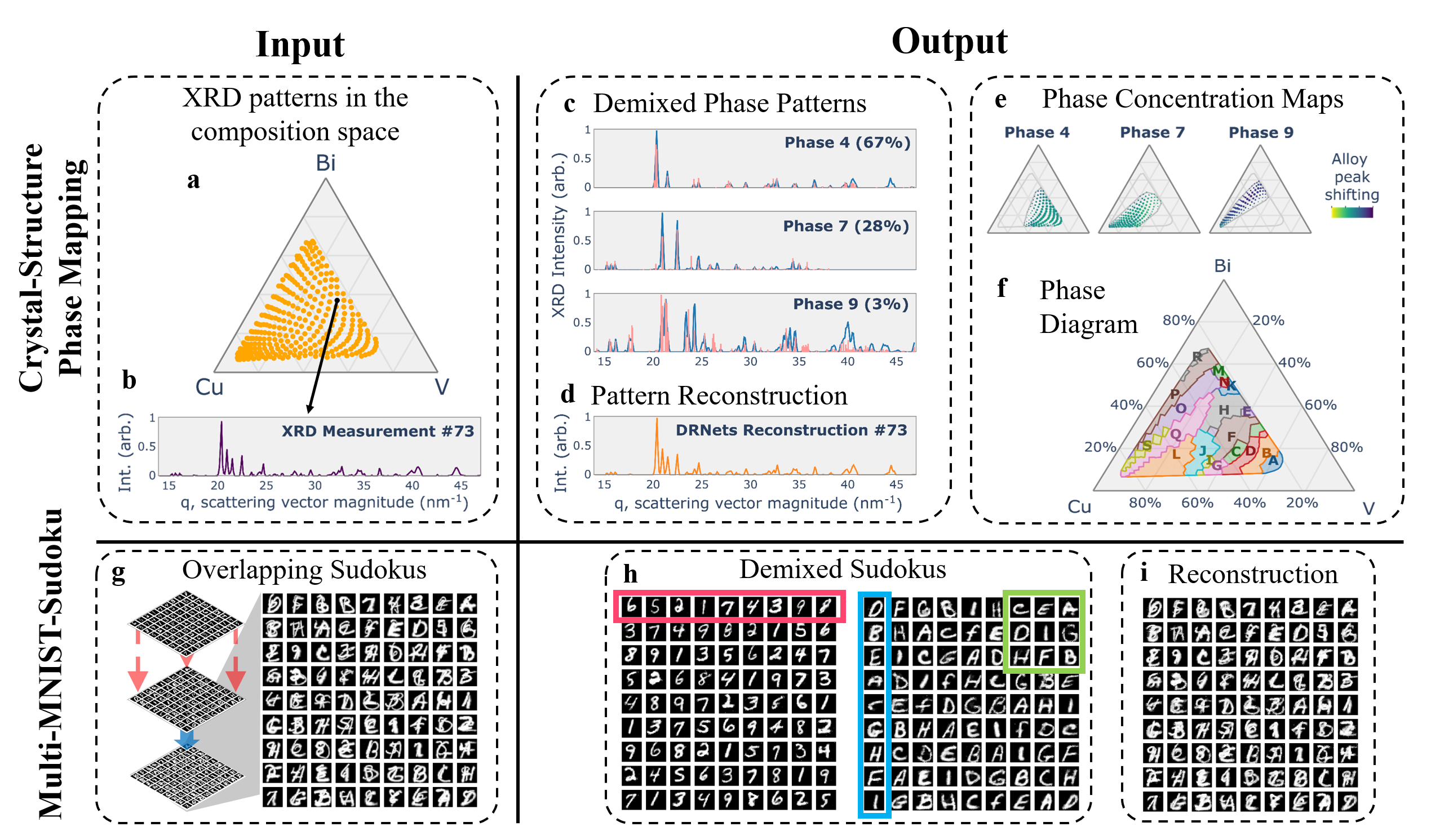} 
    %\linewidth]{FIGS/sudoku-all-7.pdf} 
\caption{
%\small{\textbf{Multi-MNIST-Sudoku (4x4) and corresponding DRNet.}}
\textbf{
 Crystal-Structure Phase Mapping and Multi-MNIST-Sudoku}  
  Phase mapping is a demixing task wherein a phase diagram is inferred from a set of XRD patterns in a materials composition space \textbf{(a)}, requiring identification of pure-phase prototypes and their composition-dependent modification. The input \textbf{(a-b)} and output \textbf{(c-f)} are illustrated for pattern \#73 where the DRNets-modified prototypes are shown as sticks in \textbf{(c)} for each demixed pattern. For each phase, DRNets output includes the composition map of activation and alloying-based modification from the prototype, shown in \textbf{(e)} for 3 phases. The composition regions corresponding to each unique combination of phases is the most salient aspect of the underlying phase diagram \textbf{(f)}. 
   In a 9x9 Sudoku, the cells in each row (red rectangle), column (blue rectangle), and any of the nine non-overlapping 3x3 boxes (green square) have all-different digits. 
   In Multi-MNIST-Sudoku, given images of mixed digit pairs, and prior
      knowledge that they form two overlapping Sudokus (\textbf{g}), the goal
      is to demix the digits into the two original Sudokus
      (\textbf{h}), closely reconstructing the original input images (\textbf{i}). 
      %See also Fig.~\ref{fig:latent_space}. 
      }
  \label{fig:sudoku}
\end{figure}

Crystal-structure phase mapping involves separating noisy mixtures of X-ray diffraction (XRD) patterns into the source XRD signals of the corresponding crystal structures, a task for which labeled training data are typically not available. Furthermore, a  valid phase diagram of the crystal structures of a given chemical system must satisfy   thermodynamics rules  (Fig.~\ref{fig:sudoku}a-f). Herein we provide a detailed description of  how to formulate phase mapping as an unsupervised pattern demixing problem and how to solve it using Deep Reasoning Networks (DRNets)~\cite{Chen2020drnets}. DRNets are a general framework for combining deep learning with constraint reasoning for incorporating scientific prior knowledge. \Carlanew{DRNets are designed with an interpretable latent space for encoding the prior-knowledge domain constraints, enabling seamless integration of constraint reasoning into neural network optimization.  
%and seamlessly integrate constraint reasoning into neural network optimization.
Constraint reasoning is a particular type of AI reasoning in which axioms and rules are expressed as constraints and the inference procedure is a search method. The axioms and rules pertaining to a given task comprise the prior knowledge needed to identify valid solutions.}
%Constraint reasoning is a particular type of AI reasoning in which axioms and rules about feasible solutions for a given task are expressed as constraints and a search method  is used as the inference procedure. }
In this manuscript,  we  show how DRNets require only a modest amount of (unlabeled) data and compensate for the limited data by exploiting and magnifying the rich scientific prior knowledge about the thermodynamic rules that govern the mixtures of crystals. %\john{DELETE[[[ with constraint reasoning seamlessly integrated  into  neural  network  optimization]]]}. 
We further  provide  insights concerning the  interpretability and scalability of  DRNets, as well as the role of data and the different  DRNets'  modules, through a series of ablation studies. DRNets make this crystal-structure  phase mapping advancement by combining  learning with constraint reasoning, emulating the analysis of expert scientists and enabling interpretation of complex systems in high-dimensional composition spaces.
%\Carlab{Herein we consider unsupervised  pattern demixing problems, which involve decomposing a mixed signal into the collection of source patterns, motivated by Crystal-Structure Phase Mapping, a core long-standing challenge in materials science. Crystal-Structure Phase Mapping involves  separating \Carlad{noisy}  mixtures of X-ray diffraction (XRD) patterns  into the source XRD signals of the corresponding crystal structures~\citep{green2017fulfilling}. 
%\Carlac{%

Given the scientific complexity of Crystal-Structure Phase Mapping, we provide  an initial intuitive explanation  of DRNets framework based on 
 Multi-MNIST-Sudoku,~\cite{Chen2020drnets} a variant of the Sudoku game that involves   demixing  two completed  overlapping hand-written Sudokus (Fig.~\ref{fig:sudoku}g-i). To demonstrate  the scalability of DRNets,  we also consider 9x9 Sudoku instances combining \textit{both digits and letters}, beyond the 4x4 Multi-MNIST-Sudoku instances involving only digits, used in the original Multi-MNIST-Sudoku variant\cite{Chen2020drnets}.  We note that,  in addition to its intuitive allure,   Sudoku represents a  logical reasoning task 
and is a computationally hard combinatorial problem~\cite{yato2003complexity}. 
Thus, Multi-MNIST-Sudoku, with hand-written digits and letters,  encapsultates a  hybrid reasoning-learning task and  provides a  tangible demonstration of  the value of %seamlessly
integrating learning and reasoning for noisy data. \Carlanew{The availability 
of ground truth data also facilitates  algorithm comparisons and ablation studies.}
%to elucidate the  role of the different components of our proposed framework.

\Carlanew{Deep Reasoning Networks (DRNets) provide a general framework that  integrates pattern recognition with   reasoning about prior knowledge.
%for unsupervised pattern demixing tasks such as crystal-structure phase mapping.  
}
%DRNets are suitable for unsupervised pattern demixing when there is rich prior knowledge about the underlying patterns. This prior knowledge can take the form of constraints that govern the patterns as well as prototypes or a generative model of the patterns. 
\Carlanew{Both Crystal-Structure Phase Mapping and Multi-MNIST-Sudoku involve identification and demixing of the component signals in mixed-signal input data, \revised{specifically crystal phases or handwritten digits and letters}. Moreover, for
%unlike with the Multi-MNIST-Sudoku, for which we can generate large-scale 
%datasets from the MNIST dataset, 
scientific tasks such as  crystal structure phase mapping, researchers generally  only have access to \Carlad{at most a few hundred (unlabeled) data samples,} %and no labeled training data, 
which greatly challenges classical data-hungry supervised deep learning models.
%Here I don't wanna mention self-supervision to trigger the reviers.
Therefore, to tackle such unsupervised demixing tasks, supervision by constraint reasoning is  \Carlad{%desired,
required 
%due to the lack of large example datasets, 
and %strongly 
supported 
%motivated 
by extensive} prior knowledge from sources ranging from fundamental \Carlad{physical} principles to the intuitive experience of scientists. More specifically, both demixing  tasks involve  two 
 types of prior knowledge: prototypes of the component signals and  rules that govern their mixtures. 
 Both demixing tasks require constraint  reasoning to interpret noisy and uncertain data, while satisfying a set of rules: thermodynamic rules,  and Sudoku rules, respectively.}
%\john{DELETE[
%\Carlab{Crystal structure phase mapping is  substantially more complex than Multi-MNIST-Sudoku. 
%%The task   easily becomes too complex for experts to solve and is a major bottleneck in high-throughput materials discovery.
%While materials science experts excel at solving simple systems,  they cannot solve complex systems, creating a major bottleneck in high-throughput materials discovery.}
%]}
\Carlanew{When considering complex data instances with multiple composition degrees of freedom and many constituent phases, crystal structure phase mapping is substantially more complex than Multi-MNIST-Sudoku and can even surpass the analytical capabilities of human experts.}

\Carlanew{Complex constraints, such as the thermodynamic rules of phase mapping, are ubiquitous in the physical sciences. Constraint satisfaction and optimization is an impactful approach for domains ranging from satisfiability to sphere packing and protein folding,\cite{rossi2006handbook,gravel_divide_2008,elser_searching_2007} and is an approach we have explored for phase mapping.\cite{lebras2011constraint} The lack of labeled data combined with the realities of experimental data, such as noise and deviations of measured patterns from their idealized prototypes, require simultaneous learning of the de-mixed signals and reasoning about their mixtures\Carlanew{, making constraint satisfaction necessary but insufficient for phase mapping solvers}.}
\Carlanew{DRNets encode complex constraints via a meaningful and interpretable latent representation coupled with a fixed generative decoder that captures the prior knowledge about the domain patterns in an end-to-end deep net framework. Furthermore, the constraint reasoning of DRNets enhances the learning of the shared parameters that govern pattern mixing across multiple (unlabeled) input instances, which in turn facilitates demixing of each pattern in the source dataset. The goal of the present work is to demonstrate the impact of this seamless integration of reasoning and learning for unsupervised pattern demixing tasks, Multi-MNIST-Sudoku as an illustrative example and ultimately crystal structure phase mapping, 
%which are solved by \Carlanewc{some} task-specific \Carlanewc{sub-component} models \Carlanewc{comprising} the same general DRNets framework.
\Carlanew{which are solved by  the same general DRNets framework customized with  task-specific \Carlanew{component} models.}
DRNets for phase mapping \Carlanew{tackle a core long standing problem in materials science,   outperforming prior methods, which is demonstrated  on a benchmark system and by solving the previously unsolved  Bi-Cu-V oxide phase diagram.} The results contribute to the broader goal of establishing DRNets as a modular end-to-end framework for tasks that require integrating pattern recognition capabilities with reasoning about prior knowledge, which are pervasive in scientific areas as diverse as biology, materials science, and medicine.}

%\Carlanew{DRNets encode complex constraints via a }
%\Carlac{
%\textit{meaningful and interpretable latent representation}  coupled with a \textit{generative decoder that captures the prior  knowledge} about \textit{the shape of the domain patterns} in an end-to-end deep net framework. Furthermore, the constraint reasoning of  DRNets  enhances  the learning of  the shared parameters concerning the way the patterns are mixed  across multiple (unlabeled)  input instances.}

%\Carlac{DRNets outperform state-of-the-art  methods on the crystal-structure phase mapping problem and Multi-MNIST-Sudoku, and even outperform human experts in disentangling complex chemical systems, such as the previously unsolved Bi-Cu-V oxide phase diagram.DRNets provide an end-to-end framework with modular  components   that can  be adapted and customized to effectively solve  a variety of unsupervised pattern demixing tasks, which are pervasive in scientific areas as diverse as biology,  materials science, and medicine.}

%% file: drnets.tex
\begin{figure}[ht]
\centering
  \includegraphics [width=1.0\linewidth]{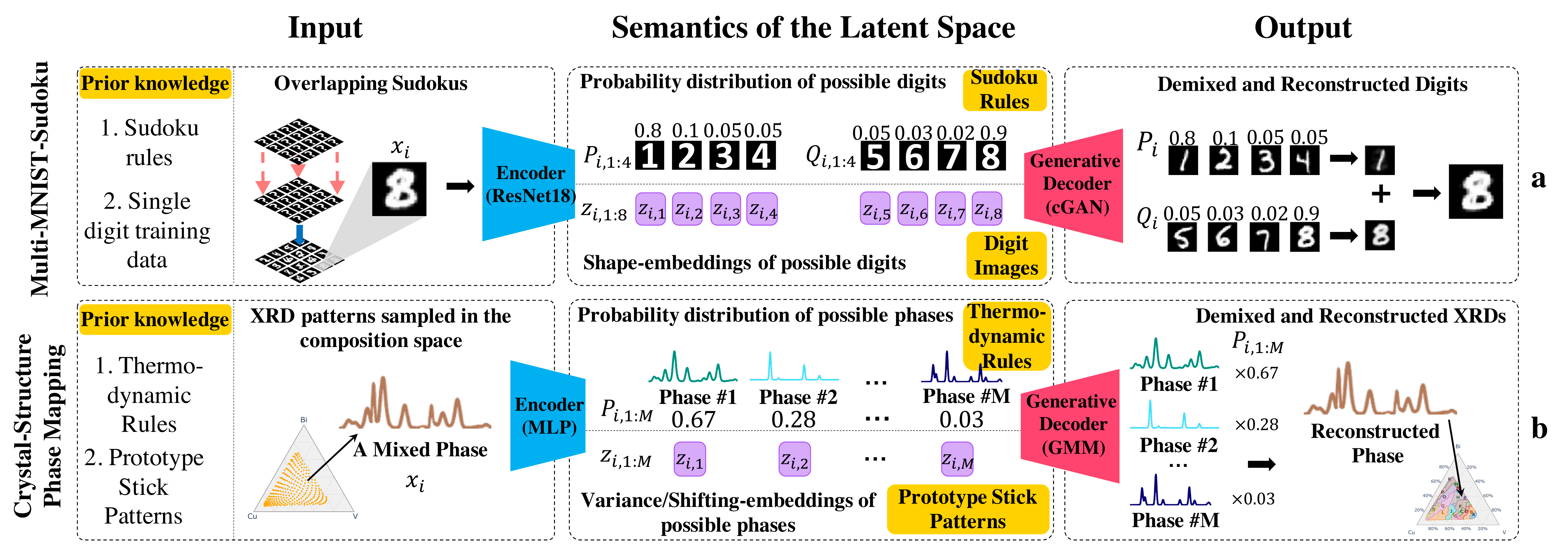}  
\caption{
%\textbf{Semantics of the latent space of DRNets for different tasks.} 
\textbf{DRNets framework and the semantics of the latent space for different tasks.} 
In the DRNets framework, an \textit{interpretable} structured latent space is key  to incorporating prior knowledge,  through the interplay of the encoder, the generative decoder (cGAN trained on single digits or a Gaussian Mixture Model (GMM) based on prototype stick patterns), and the reasoning constraints (Sudoku rules or thermodynamic rules).
%the reasoning constraints and the generative decoder. 
%digit and ICDD crystal-structure phase probabilities and shape parameters for the Sudoku and Crystal-structure phase mapping applications, and species preferences and interactions (shared co-variance matrix) for the join species distribution modelling.
%
\textbf{a.} In \textbf{Multi-MNIST-Sudoku}, DRNets encode the input overlapping digits $x_i$ into $P_{i,1:4}$, $Q_{i,1:4}$ and $z_{i,1:8}$, which denote the probability distribution and the shape embedding of possible digits (1-8).  
The generative-decoder (cGAN) uses $z_{i,1:8}$ to generate the demixed hand-written digits and reconstruct the original input with the expected overlapping image using $P_{i,1:4}$ and $Q_{i,1:4}$.
\textbf{b.} In \textbf{crystal-structure phase mapping}, DRNets encode the input XRD pattern $x_i$ into $P_{i,1:M}$ and $z_{i,1:M}$, which denote the probability distribution and the variance/shifting-embedding of $M$ possible phases.
The generative-decoder (GMM) uses $z_{i,1:M}$ to generate the decomposed phases and reconstruct the original XRD using the phase probability distribution $P_{i,1:M}$.
 }
 \label{fig:latent_space}
\end{figure}
\section*{\small{\Carla{DRNets framework}}}
At a high level, the goal of unsupervised pattern demixing of \revised{crystal structures or digits and letters }
%demixing and completion
 is  \Carlab{to infer the base patterns underlying the mixtures observed in unlabeled data.\cite{footnote}}
 %underlying the mixture from unlabeled data.
 %to infer patterns from unlabeled data.\cite{footnote}
 The demixing  task is therefore to invert \Carla{the pattern mixing  processes from the data,
%the mixing pattern generative model,
i.e., the generative processes for each pattern  %for which we have a generative model, 
and the way the patterns are  combined.}
% that is encapsulated in the reconstruction loss function and reasoning loss,  
%, typically by minimizing a  reconstruction loss of the input data.  
However, often unlabeled data do not  provide a strong \Carlab{enough} pattern signal, \Carlab{motivating reasoning about prior knowledge.} %, a challenge for standard machine learning approaches.}
%.  In the absence of labeled datasets, additional information  can come from   prior knowledge concerning domain rules or constraints (e.g., Sudoku rukes or thermodynamic rules) or in the form of a fixed pre-trained generative model or a parametric generative model of the patterns, which can be  obtained for example from a  theoretical model of pure patterns based on prior knowledge (e.g., pre-trained digit prototypes or idealized \textit{stick patterns} of known crystal structures).  
\Carla{
 DRNets enhance standard  unsupervised pattern discovery
 %learning 
 approaches with} \Carlanew{prior knowledge about the} constraints that govern the  patterns, via  constraint reasoning, and prior knowledge about the patterns' \textit{shape,}  via a \Carlanew{fixed} generative decoder (see Fig.~\ref{fig:latent_space}).  
 \Carlab{
 More specifically, 
 DRNets 
  combine deep learning with constraint reasoning in an \textit{end-to-end encoder-generative-decoder framework,} and 
 formulate unsupervised pattern discovery
 %learning
 as a \textit{data-driven constrained optimization problem} that  \textit{(i)} minimizes  a reconstruction loss of the input data, such that \textit{(ii)} the inferred patterns adhere to a  given generative model  and \textit{(iii)} satisfy  domain constraints. \Carlab{See mathematical details about DRNets' problem formulation flow in  Extended Data Fig.~\ref{fig:DRNet-flow} and %\john{Methods}
 \hyperref[subsec:formulation]{Methods}.
 }
 }
 
 \Carla{While standard machine learning approaches can easily handle \textit{(i)}, enforcing \textit{(ii)} and \textit{(iii)} is challenging and emblematic of the limitations of traditional deep learning. 
 \Carlab{The standard approach for incorporating domain knowledge into a deep net architecture is to add terms to the loss function such as various types of sparsity  constraints. However, 
 %the kinds of constraints
 we need to encode more complex constraints, such as combinatorial constraints to express 
valid Sudoku solutions or thermodynamic rules.
 For example,  a digit cannot appear more than once in a row or an X-ray diffraction pattern cannot be explained by more than 3 
 \Carlanew{prototype phases,}
 %crystal phases (or patterns) 
 or 2 
 \Carlanew{prototype phases}
 %crystal phases 
 if there is alloying. \Carlad{\textit{The challenge of our demixing tasks is that the domain rules capturing prior knowledge involve variables   that we do not have direct access to in our problem formulation. In fact, discovering those variables and their values is part of the interpretation task that we are trying to solve.}} The challenge is further complicated by the fact that we are operating in an unsupervised setting (no labeled training data). 
 \Carlanew{In supervised learning, different strategies have been exploited to incorporate prior knowledge, ranging from placing constraints on the output variables of the deep net  to hybrid approaches  interleaving symbolic and neural processing.~\cite{garcez2020neurosymbolic}}
 %In supervised learning, prior knowledge can often be captured by placing constraints on the output variables of the \Carlanew{deep network~\cite{garcez2020neurosymbolic}.} 
 However, in standard unsupervised deep learning approaches, e.g., neural networks autoencoders, the latent space is generally uninterpretable and therefore does not provide a means to express the domain constraints or \Carlanew{enforce}
 %enforcing 
 that the latent patterns conform to the generative model.}
 %In standard unsupervised  deep learning approaches, e.g., neural network autoencoders,  the latent space is uninterpretable and therefore it does not provide a  means to express the domain constraints or enforcing that the latent patterns conform to the generative model. 
 %Other unsupervised approaches such as latent based models suffer from the same limitation.  
 Furthermore, domain  rules are often  captured by combinatorial constraints that are not differentiable, and therefore cannot be easily embedded in a deep learning framework. The strategy in DRNets to  overcome these challenges  is  to 
 \Carlanew{ \textit{(1)}  specify \textit{an intended semantics} for the  latent space, which means that the latent space is constructed using variables that have a specific interpretation that can be used in the formulation of the domain rules. For example, in the Sudoku domain, we will introduce a latent variable for each possible digit that gives the probability of that digit being present in the cell associated with the input image. We can now use these variables in constraints on the allowed combinations of digits }
 %\textit{(1)} specify  \textit{an intended semantics} to the encoder's latent space,
% \Carlanew{which means that the latent space  is constructed with variables with %using variables that have 
%an intended} %physical 
%\Carlanew{representational meaning}  %provides the basis for formulating 
%\Carlanew{ for the encoding of the domain rules.}
% \john{DELETE[[[\Carlab{using variables} that enable the encoding of the domain  rules and the parameters required for the generative decoder]]]}
 (see Fig.\  \ref{fig:latent_space}).  \Carlab{As we will show, these latent variables take on the desired semantics using %\john{DELETE[[[only a surprisingly]]] }
 a small set of unlabeled examples combined with the encoded  domain constraints;}  \textit{(2)}  express the domain rules that control the encoding of the latent space  in a form amenable to continuous optimization using {\textit{entropy-based continuous relaxations;}}  
 \textit{(3)} employ \textit{an optimization formulation  whose objective  balances
 the dual needs of minimizing a  reconstruction loss of the  input data  and a reasoning loss that captures the domain  constraints}
 (local constraints involving a single data point or global constraints involving 
 %several 
 \Carlanew{many} data points);
  \textit{(4)}  \textit{use a data-driven approach  \Carlab{to jointly solve} multiple related (unlabeled) data instances}; and  \textit{(5)} solve the data-driven constrained optimization problem  using  \textit{constraint-aware stochastic gradient descent}, 
%(See \hyperref[subsec:formulation]{Methods} and  \hyperref[SI:SGD]{SI~\ref{SI:SGD}}), 
a variant of stochastic gradient descent  \Carlab{developed  for DRNets} that batches together  data points involved in  the same constraints  and \textit{is aware} of the constraints, automatically adjusting the weights of the constraints as a function of their satisfiability. }

%Di: Looks good to me! 
%\Carlanewc{In DRNets, two intertwined processes %intended 
 %produce the encoder's weights that return %the correct parameters 
 %the %correct 
%parameters of the interpretable latent space variables seamless combine  learning and reasoning:}
\Carlanew{In DRNets, learning is data driven and reasoning is knowledge driven. Two intertwined processes  combine learning and reasoning to discover the values for the encoder's parameters that provide the best interpretation for the interpretable latent space, given the data and prior knowledge:} %variables' parameters: }
\Carlanew{input pattern reconstruction (digits and crystal phases) in conjunction with reasoning about the domain rules (Sudoku and thermodynamic rules).  The input pattern reconstruction is performed  through the reconstruction loss, with %"supervision" from %prior knowledge concerning 
guidance from 
prototypical  domain patterns (single digits and crystal phases) provided via a  fixed generative decoder. The reasoning is performed with "self-supervision" from  the domain rules prior knowledge, encoded as constraints using the interpretable latent space variables and added as entropy-based continuous functions to the loss via Lagrangean relaxation. The reconstruction and reasoning loss constrain the encoding of the latent space to adhere to the domain patterns and rules.
}\Carla{We refer to DRNets' formulation as  \textit{data-driven constrained optimization} to highlight \Carlad{that even though} \Carlanew{we design an interpretable latent-space, its semantics} \Carlanew{are ultimately determined by the quality and quantity of the (unlabeled) data as well as the prior knowledge available, which are critical for conditioning the optimization process to discover the underlying patterns and pattern mixing process across instances.} Since often the pattern mixing function is  not invertible, %\john{DELETE[without enough (unlabeled) data]}, 
the  optimization process in DRNets is  further conditioned  by learning the shared mixing process parameters across instances \Carlad{(e.g. multiple Sudoku instances or  %\john{DELETE[libraries of]} 
related X-ray diffraction patterns).} In contrast, standard direct optimization typically considers one instance at a time. 
\Carlab{Interestingly, as our ablation studies show, while multiple (unlabeled) data are required by DRNets to uncover the pattern mixing process, the amount of (unlabeled) data required is considerably more modest than in standard supervised deep learning settings (see Fig.~\ref{fig:down-scale-PM} and Extended Data Fig.~\ref{fig:down-scale-mnist}).}
%(hundreds instead of thousands of instances). }
%In other words, the generative decoder  maps from an interpretable constrained latent space  to the  input data and the optimization task is to infer the  semantics of the latent space based on the data. 
%Furthermore, while in standard constraint optimization problems  the parameters of the  model  (e.g., coefficients of objective function and constraints) are obtained \textit{a priori} from the data, in the DRNEts framework (some of)  such  parameters are optimized  concomitantly  with pattern inference from the data. 
%In the next sections we apply the DRNets framework to Sudoku demixing and crystal-structure phase mapping.
%, a demixing problem for inferring crystal structures from data, a core problem in materials discovery. 
%We provide several ablation studies to elucidate the role of the data and the different components of the DRNets framework.
Further details about DRNets' problem formulation flow, constraint relaxations, 
and DRNets' algorithms are given in  Extended Data Fig.~\ref{fig:DRNet-flow},
%\john{Methods and Supplementary Methods}
%a,b, 
\hyperref[subsec:formulation]{Methods}, and \hyperref[sec:si]{Supplementary Methods}.
 }
 %, and \hyperref[SI:SGD]{SI~\ref{SI:SGD}}.

%% file: main-sudoku.tex
\section*{\small{Sudoku: demixing  handwritten digits and letters}}

%\john{DELETE[DRNets are inspired and motivated by 
%scientific tasks,
%such as \Carla{crystal-structure phase mapping in materials discovery.}
%The underlying scientific domain is complex and involves previously unsolved problems, motivating our introduction of DRNets using  Multi-MNIST-Sudoku.]}
%a demixing variant of the popular Sudoku game where ground truth data facilitates  comparison to supervised learning algorithms. 
\begin{comment}|
\Carla{In addition to its intuitive allure,   Sudoku is a prototypical logical reasoning task 
%as it requires inferring missing digits from the given hints 
and it is also a prototypical computationally hard combinatorial problem that belongs 
to the so-called class NP-complete. Essentially this means that while we can easily verify whether a solution satisfies the Sudoku rules in polynomial time (class NP), generating its solution is exponentially harder in the worst case, as the number of possible ways of filling in the empty Sudoku cells  grows exponentially with the instance size (4x4, 9x9, etc). The Sudoku completion task can be reduced to general  propositional logical reasoning or satisfiability (SAT)  and vice-versa. Crystal-phase mapping is also a demixing  computationally hard combinatorial problem~\cite{lebras2011constraint}.}
%The underlying scientific domains are quite complex. We therefore  first introduce DRNets with an illustrative game, 
%To provide an intuitive overview of DRNets, we commence with an
%challenging illustrative game,
\end{comment}
Multi-MNIST-Sudoku  consists of demixing digits from two \Carlab{completed} overlapping hand-written Sudokus
while satisfying Sudoku rules (Fig.~\ref{fig:sudoku}a-c).
%\john{, where the knowledge that each cell contains 2 overlapped digits combined with the Sudoku rules comprises the prior knowledge (Fig.~\ref{fig:sudoku}f).}
\Carlanew{The prior knowledge comprises the information that a set of images forms two overlapping handwritten Sudokus; each image corresponds to a Sudoku cell with two overlapping handwritten  digits/letters; and the Sudoku rules (Fig.~\ref{fig:sudoku}f).}
%Multi-MNIST-Sudoku, which  consists of demixing digits from two overlapping hand-written Sudokus
%while satisfying Sudoku rules (Fig.~\ref{fig:sudoku}a-c). Multi-MNIST-Sudoku provides an accessible  analogy  to scientific discovery problems such as crystal-structure phase mapping.
%Both tasks require demixing signals under constraints, with phase mapping having thermodynamic constraints that are more complex than Sudoku rules as well as a scale of data beyond that of Multi-MNIST-Sudoku.
%which is  also a  demixing task, though substantially more complex than Multi-MNIST-Sudoku, both in terms of its constraint set (thermodynamic rules) and scale. 
%Multi-MNIST-Sudoku also 
%serves as an illustrative benchmark for DRNets 
%due 
%due to
%the availability of
%ground truth data.
Humans tackle Multi-MNIST-Sudoku by interleaving vision
clues with reasoning about Sudoku rules, which is emulated by DRNets
as illustrated in Fig.~\ref{fig:sudoku}d. 
%Analogously to human reasoning,
%\revised{%In an artificial 
%reasoning system  an inference procedure derives what follows from an initial set of axioms and rules.}
%In an artificial 
%reasoning system  an inference procedure derives what follows from an initial set of axioms and rules.
%\john{MOVE THIS TO INTRO[[[Constraint reasoning is a particular type of AI reasoning in which axioms and
%rules are expressed as constraints and the inference procedure is a search method.]]]}
DRNets combine deep learning with constraint reasoning and optimization to reason about Sudoku rules.
In Multi-MNIST-Sudoku DRNets, a deep neural network
encodes a structured latent representation of the input (digit images), which captures the probabilities and shapes of the possible digits under Sudoku constraints enforced by the reasoning module (Fig.\ref{fig:latent_space}a). The reasoning module comprises the reasoning constraints as well as batch and constraint weight controllers.  
A \Carlanew{fixed conditional generative adversarial network (cGAN) pre-trained on single digits, incorporating prior single digit prototype knowledge,} decodes (generates) the individual digit images
from their structured latent representation \revised{along with their probabilities to reconstruct the mixed input image}.
The overall objective
function of DRNets combines responses from the generative decoder 
and the reasoning constraints,  
and is optimized using constraint-aware
stochastic gradient descent.
%\revised{
%Note that Sudoku constraints are discrete in nature and therefore are not naturally captured by a differentiable framework.} 
We apply  entropy-based probabilistic 
continuous relaxations %that use probabilistic modelling 
to encode
discrete constraints, such as Sudoku rules, which can  be
seamlessly incorporated into the objective function. 
%\revised{We highlight the importance of DRNets'  structured  latent space  with an intended interpretation, a   key idea underlying  DRNets' framework, which is quite different from standard deep learning approaches, in which the latent space does not have a well defined semantics. The  \textit{interpretability} of the latent space  allows DRNets to encode it by tightly interweaving low level data driven information, pixels and digits, via  the  generative decoder,  and higher level concepts such as  Sudoku rules, via the constraint reasoning module. This tight coupling of the responses from the decoder with the responses from the reasoning module via the semantics of the latent code  is also key to increasing the robustness of the solutions.}
\textit{A key distinguishing feature of DRNets is an \textit{interpretable latent space} with semantics emerging  by the coupling of the encoder, the generative decoder, and reasoning module,} in contrast to standard deep learning methods in which the latent space lacks semantics.
%\revised{We also note that latent representations of input data are a tenet of deep learning methods, although in traditional methods the latent space lacks semantics. A key innovation of DRNets is an \textit{interpretable latent space } with semantics emerging  by the coupling of the generative decoder and reasoning module.} 
Further details about DRNets’ components,  latent space semantics, problem formulation, and algorithms are given in 
Fig.~\ref{fig:latent_space}, Extended Data Fig.~\ref{fig:drnets_mnist}, and
%Extended Data (Fig.~\ref{fig:DRNet-flow} and ~\ref{fig:DRNet-latent})
%\john{Methods.}
 \hyperref[subsec:formulation]{Methods}. %\hyperref[SI:InformalSudoku]{SI~\ref{SI:InformalSudoku}}. 
%

%%%%%%%%%%%%%%%%%%%%%%%

To evaluate DRNets, we generated 32x32 images of overlapping digits/letters
from 
the test set of 
MNIST \cite{lecun1998gradient} and EMNIST~\cite{cohen1702emnist}, such that every $n^2$ ($n$ is 4 or 9) images 
%\Di{digit/letter images}
form two n-by-n, overlapping Sudokus 
%\john{(see Supplementary Methods).}
(see \hyperref[sec:si]{Supplementary Methods}). 
For the 9x9 case, to distinguish the two overlapping Sudokus, we used letters A-I from EMNIST 
%as the digits 1-9 
for the second Sudoku. 
For ease of presentation, we refer to these letters as "digits" in the following content.
%we used digits 1-4 (digits 1-9) and digits 5-8 (letters A-I) for the 4x4 (9x9) two overlapping  Sudokus, respectively.
DRNets are unsupervised  and therefore do not train on labeled mixed digit images;
DRNets only have access to single hand-written digits from the MNIST/EMNIST training
dataset, \Carlanew{which are used to pre-train the generative decoder (cGAN). }
%We compare DRNets’ performance against the  
\Carlanew{DRNets significantly outperform state-of-the-art supervised
MNIST demixing models CapsuleNet~\cite{sabour2017dynamic} and ResNet~\cite{he2016deep}, which in contrast to DRNets 
%more informative 
are supervised by training on labeled mixed
%mixed-digit
images produced by overlapping
digits from the MNIST/EMNIST training dataset  (Extended Data Fig.~\ref{table:mnist}). }
%DRNets recovered all digits perfectly, outperforming CapsuleNet and ResNet, 
%\john{DELETE[DRNets significantly outperform CapsuleNet and ResNet, 
%even when these supervised systems are coupled with local search to incorporate Sudoku rules.
%This is particular true when going from 4-by-4 to  9-by-9 Multi-MNIST-Sudoku
%instances:  DRNets' digit and Sudoku accuracy for 9-by-9 instances is close to  100\%, while ResNet's Sudoku accuracy is around 35\% and 80\% when boosted by local search. Also, while for the 4-by-4 we could improve the Sudoku accuracy of the supervised  methods by explicitly considering all the possible Sudoku solutions and selecting the most likely, such a strategy is out of the question for 9-by-9 instances, given that there are around $6.67\times10^{21}$ possible Sudoku solutions, which highlights the combinatorial nature of this problem.]}

\Carlanew{The intuitive nature of the Sudoku task provides an opportunity to gain insights regarding DRNets' structure and performance via ablation studies. For example, }
%Another  ablation study shows that  
if we replace the \Carlanew{fixed}  cGAN with a (weaker) standard  learnable decoder, without prior knowledge about single digits, the optimization process can no longer find the right semantics for the latent space.
%, even tough the discovered "nonsensical digits" still follow the Sudoku rules. 
Moreover, removing the reasoning module also deteriorates the digit accuracy, in addition to the Sudoku accuracy (Extended Data Fig.~\ref{table:mnist}
and 
\hyperref[sec:note]{Supplementary Note}).
%
%, \hyperref[sec:methods]{Methods}
%and \hyperref[sec:si]{SI}).
\Carlab{The final} \Carla{ablation study shows the importance of the data-driven learning of the shared parameters  of the demixing task across multiple  9x9 Multi-MNIST-Sudoku instances. %, which helps model convergence. 
Nevertheless,  DRNets  can reach 99\% accuracy with only 100 (unlabeled) 9x9 Multi-MNIST-Sudoku instances (Extended data Fig.~\ref{fig:down-scale-mnist}), a considerable smaller amount of data compared to standard deep learning approaches.
%Interestingly, total optimization time also decreases with increase of dataset (1.5 hours for 100 vs.\ 0.87 hours for 1,000) \Di{(This running time difference is mainly due to the batch size difference (a larger batch is more GPU efficient.) I don't really think we should talk about it. When I run DRNets with 100 instances, I have to use a smaller batch size to make it perform better.)}.  
More generally, DRNets accomplish superior performance through  {\textit{self-supervision by reasoning}} about 
Sudoku rules, resulting in a better performance for digit accuracy and considerably better for Sudoku accuracy than the supervised systems. }% (Extended Data Table \ref{table:mnist}). 
The broader implication is that the reasoning component enables DRNets to compensate for the lack of training data, which considerably broadens the purview of AI to problems that have a rich prior knowledge but may lack sufficient training data for traditional deep learning.
%See more details in \hyperref[SI:Sudoku]{ SI~\ref{SI:Sudoku}}.

%SI:Informal_Sudoku

%%%%

%% file: main-phase-mapping.tex
\section*{\small{Crystal-structure phase mapping: demixing X-Ray diffraction patterns}}
\begin{figure}
\centering
  \includegraphics [width=0.95
  \linewidth]{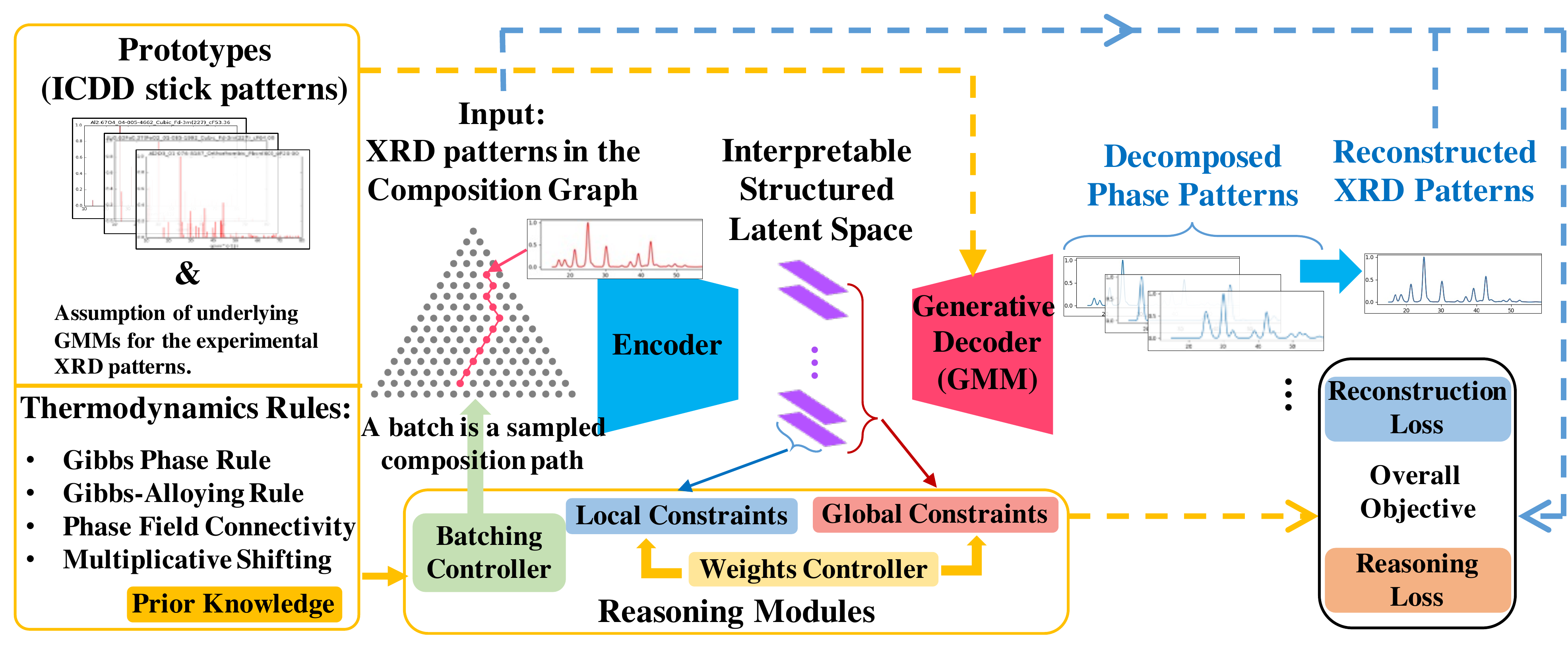}
    %\linewidth]{FIGS/pm-new-fig.png} 
  \caption{
  \textbf{
%   DRNets crystal structure phase mapping 
%   %for \Di{chemical systems}
%   %the Bi-Cu-V oxide system 
%   and comparison with other unsupervised approaches for the Bi-Cu-V oxide and Al-Li-Fe oxide systems.
 %Crystal-Structure Phase Mapping and corresponding DRNet
 DRNets for Crystal-Structure Phase Mapping
 } 
%   Phase mapping is a demixing task wherein a phase diagram \textbf{(f)} is inferred from a set of XRD patterns in a materials composition space \textbf{(a)}, requiring identification of pure-phase prototypes and their composition-dependent modification. The input \textbf{(a-b)} and output \textbf{(c-f)} are illustrated for pattern \#73 where the DRNets-modified prototypes are shown as sticks in \textbf{(c)} for each demixed pattern. For each phase, DRNets output includes the composition map of activation and alloying-based modification from the prototype, shown in \textbf{(e)} for 3 phases. The composition regions corresponding to each unique combination of phases is the most salient aspect of the underlying phase diagram \textbf{(f)}. 
  %
  %
  \Carla{
  %\textbf{g.}
  DRNets perform end-to-end deep reasoning by encoding a latent space that is used by a generative decoder to reconstruct the XRD measurements. 
The input is the XRD patterns, each resulting from a mixture of phases, and the output is the decomposed pure phase patterns and the reconstructed mixture.
    The encoder is composed of four
    3-layer-fully-connected networks. The structured latent encoding is constrained to adhere to thermodynamic rules by the reasoning module.
    Prior knowledge also includes prototype stick patterns, 
    which are used by the generative decoder, a Gaussian mixture model, to generate the corresponding possible phase patterns in the reconstructed XRD measurement.
    An overall objective combines responses from the generative decoder, for pattern reconstruction, and the reasoning module, for applying thermodynamic rules, which is optimized using constraint-aware stochastic gradient descent.
    }
  }
  \label{fig:bi-cu-v-o}
\end{figure}

\begin{figure}
\centering
  \includegraphics [width=0.98\linewidth]
  %{FIGS/bi-cu-v-drnets-all-phase-fields.png} 
  {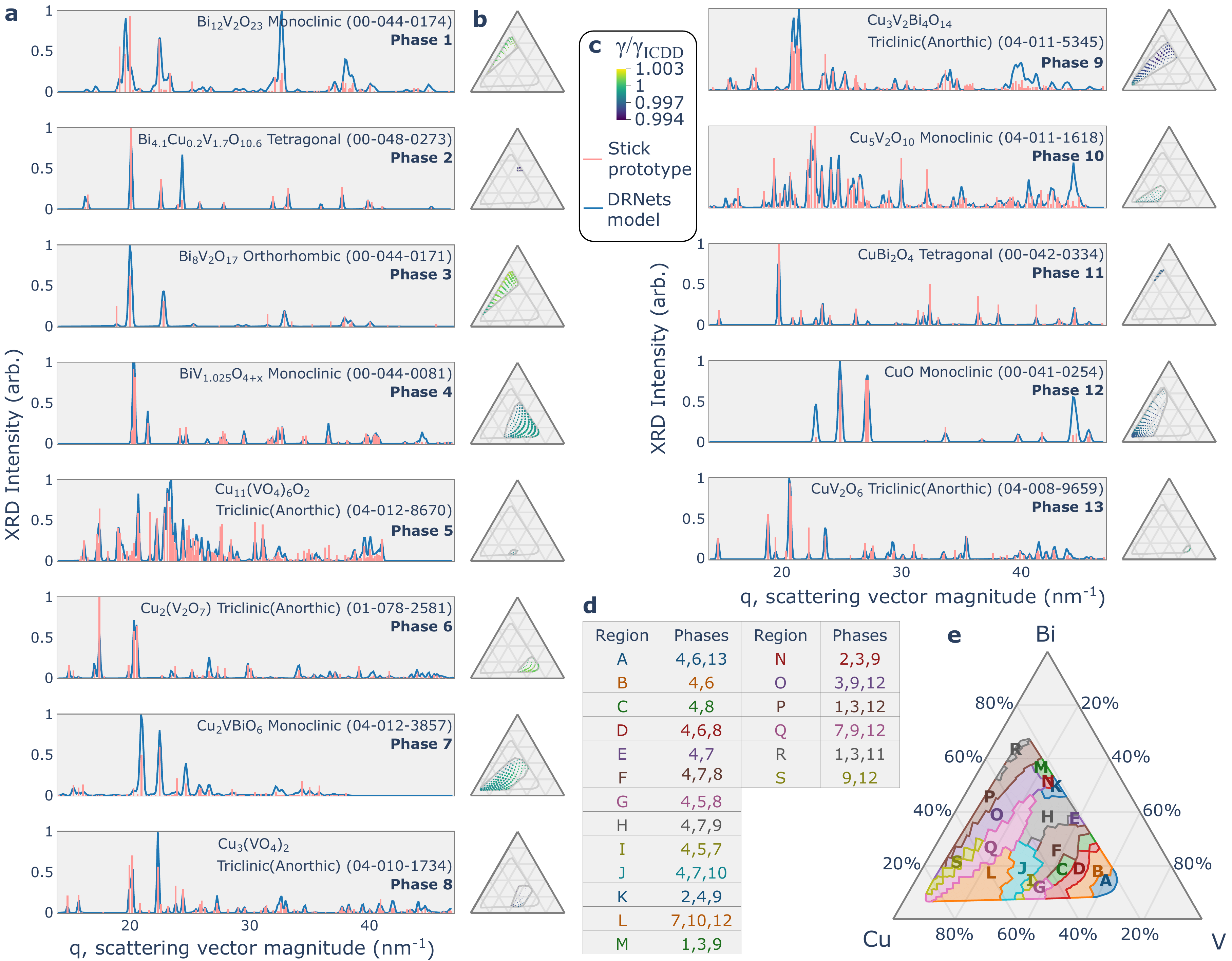}
  \caption{\small\textbf{DRNets'
  solution for the Bi-Cu-V oxide system.}  \textbf{ a.} The \Carlanew{13} demixed
  crystal phases for the 307 XRD measurements of the Bi-Cu-V oxide
  system (each plot includes the signal for the recognized phase and the
  corresponding ICDD stick pattern). \textbf{b.} DRNets' phase concentration maps for \Carlanew{each of the phases, where point}  sizes are
  proportional to their estimated phase concentrations and heatmap
  denotes estimated shifting (alloying). \Carlanew{ \textbf{c.} Shows the universal legend for the 13 phases in \textbf{a} and \textbf{b}, where $\gamma$ is the average lattice constant.} \Carlanew{\textbf{d.} Table of all phase mixtures in the DRNets solution. \textbf{e.} DRNets' crystal phase map for the Bi-Cu-V-O system with phase fields labeled according to \textbf{d.}}}   \label{fig:extra-bi-cu-v-o-drnet}
\end{figure}

Scientific discovery comprises a range of problems where
{\textit{self-supervision by reasoning}} is strongly desired  due to
the lack of large example datasets, and strongly motivated by
extensive prior knowledge,
%from sources ranging 
from fundamental
principles to the intuitive experience of
scientists. 
In materials science, phase mapping is the problem of  inferring the individual phases, i.e., crystal structures, from \Carlanew{a collection of} X-ray diffraction (XRD) patterns (Fig.\ref{fig:sudoku}a-f),  a major bottleneck in research
 due to the substantial expert analysis required for
generating meaningful solutions~\cite{ludwig2019discovery,green2017fulfilling,kusne2015high,stanev2018unsupervised,gomes2019crystal}.
As a demixing problem, phase mapping parallels the high level structure of Multi-MNIST-Sudoku where instead of overlapping handwritten digits, an XRD pattern contains a mixture of signals from so-called “pure” phases, and the solution includes demixed signals from each XRD pattern with rules based on a collection of input XRD patterns, as illustrated in  Fig.~\ref{fig:bi-cu-v-o}. The prototypes in phase mapping are \textit{stick patterns} that provide the locations and intensities of peaks in XRD patterns for each known phase (See Fig.~\ref{fig:bi-cu-v-o} and Fig.~\ref{fig:extra-bi-cu-v-o-drnet}), and this set of peaks comprises the entirety of the XRD signal for a single crystal structure, i.e., a pure-phase pattern.
Computationally, phase mapping is an
NP-hard problem~\cite{lebras2011constraint} whose sheer number of possible combinations of prototypes in each XRD pattern 
%base patterns and activations 
grows exponentially with data size,
($\sim $ 300 XRD patterns and $\sim$ 200 prototypes),
rendering
traditional methods computationally infeasible. 
Supervised methods \cite{park_classification_2017,sun_accelerated_2019,Ovideo2019,lee2020deep,bunn_generalized_2015} for phase identification in an XRD pattern and unsupervised methods for phase mapping\cite{stanev2018unsupervised,gomes2019crystal,rossouw2015multicomponent} have been developed and perform well on some datasets, especially when input patterns are akin to simple mixtures of prototypes with mutually distinguishable features. 
Ternary composition spaces, i.e., mixtures of 3 elements from the periodic table, have great scientific value for discovery of materials with desired properties that is concomitant with complex phase behavior in which different compositions (proportions of the elements) form many unique mixtures of the prototype phases.
%The scientific value of phase mapping is centralized capitalizedin combinations of 3 or more elements from the periodic table that exhibit complex phase behavior in which different compositions (proportions of the elements) form many unique mixtures of the prototype phases.
The complexity is compounded by phenomena such as alloying, wherein each composition’s XRD pattern may contain unique variants of the prototypes. The most typical alloying-based variation of a prototype includes altered peak intensities and systematically-shifted peak positions. \Carlanew{Peak intensities can also vary from the prototypes due to various ways in which the materials and experiment conditions vary from those used to generate the prototypes.} When the variants of prototypes contain strongly overlapped signals \Carlanew{with unknown relative peak intensities compared to the prototypes}, %(analogous to overlapped handwriting), 
phase mapping can only be solved (by humans or AI) through reasoning about prior knowledge based on thermodynamics. These “rules” are more nuanced than those of Sudoku and are described in detail in
\hyperref[sec:si]{Supplementary Methods}.
%and
%\john{Supplementary Methods.}
%\hyperref[SI:phase_mapping_relaxations]{SI~\ref{SI:phase_mapping_relaxations}}.
Briefly, when combining 3 elements, the number of phases that can appear in an input pattern is at most 3, and is at most 2 if the composition is in a region that exhibits alloying. This latter rule requires consideration of the 
composition graph of the input XRD patterns,
%\John{variation in each phase's signal between XRD patterns from neighboring compositions, further exacerbating the combinatorics of the problem}
which is also used to enforce a rule wherein each set of prototypes can only appear in compositions that are connected in the graph. 
%%%%%%%%

State-of-the-art approaches for phase mapping are unsupervised methods based on  matrix factorization, and have  incorporated thermodynamic rules to different extents, including analysis of the composition graph~\cite{kusne2015high},
integration of demixing with clustering~\cite{stanev2018unsupervised}, 
and recent work \Carlanew{that interleaves  matrix factorization with constraint optimization to enforce  all the thermodynamic rules \cite{gomes2019crystal}. Nevertheless, these  approaches  only use known prototype patterns to post-process demixing results,
%as demixing post-processing,
resulting in a more ill-conditioned demixing.} In contrast, \Carlanew{DRNets provide  the first framework that  integrates enforcement of thermodynamic rules with reasoning about 
the prototypes,  solving previously unsolvable phase mapping problems.} %(Fig.~\ref{fig:bi-cu-v-o}f-h).
%More specifically, 
\begin{comment}
The DRNets for crystal-structure phase-mapping (Fig.~\ref{fig:bi-cu-v-o}) 
seamlessly integrate demixing and reconstruction of XRD patterns by coupling 
\Carlanew{an encoder with four 3-layer-fully-connected neural networks,}
%a four 3-layer-fully-connected neural network encoder, 
which produces a two-part  structured latent space (Fig.~\ref{fig:latent_space}b), to a generative Gaussian Mixture model that 
incorporates prior knowledge by generating XRD patterns based on mixtures of modified versions of prototypes. 
\end{comment}
The DRNets for crystal-structure phase-mapping (Fig.~\ref{fig:bi-cu-v-o}) 
\Carlanew{seamlessly integrate demixing and reconstruction of XRD patterns 
by coupling 
\Carlanew{an encoder with four 3-layer-fully-connected neural networks,}
%a four 3-layer-fully-connected neural network encoder, 
which produces a two-part  structured latent space (Fig.~\ref{fig:latent_space}b), to a generative Gaussian Mixture model that 
incorporates prior knowledge about prototype  phases, and  by generating XRD patterns based on mixtures of modified versions of prototypes.} 
\Carlanew{The modifications include peak intensity modulation and alloying-based peak shifting; the semantic representation of the prototype-modification parameters in the latent space enables DRNets to learn their optimal values under guidance from priors that are parameterized by prior knowledge of the maximum extent of prototype modification. The latent variables also enable expression of thermodynamic rules with entropy-based functions, which are imposed }
%The latent space features the semantics necessary to estimate the probabilities of alterations to the prototypes while imposing the thermodynamic rules, which are encoded in entropy-based functions and applied 
with a batching sampling strategy to tackle the combinatorics of all-sample thermodynamic constraints  (Fig.~\ref{fig:bi-cu-v-o}).
DRNets are 
optimized
with the hybrid objective of reconstructing
measured patterns and enforcing thermodynamic rules, as detailed
in %\john{Methods.}
\hyperref[sec:methods]{Methods}.
%%%%

\begin{figure}[ht]
\centering
  \includegraphics [width=0.95
  \linewidth]{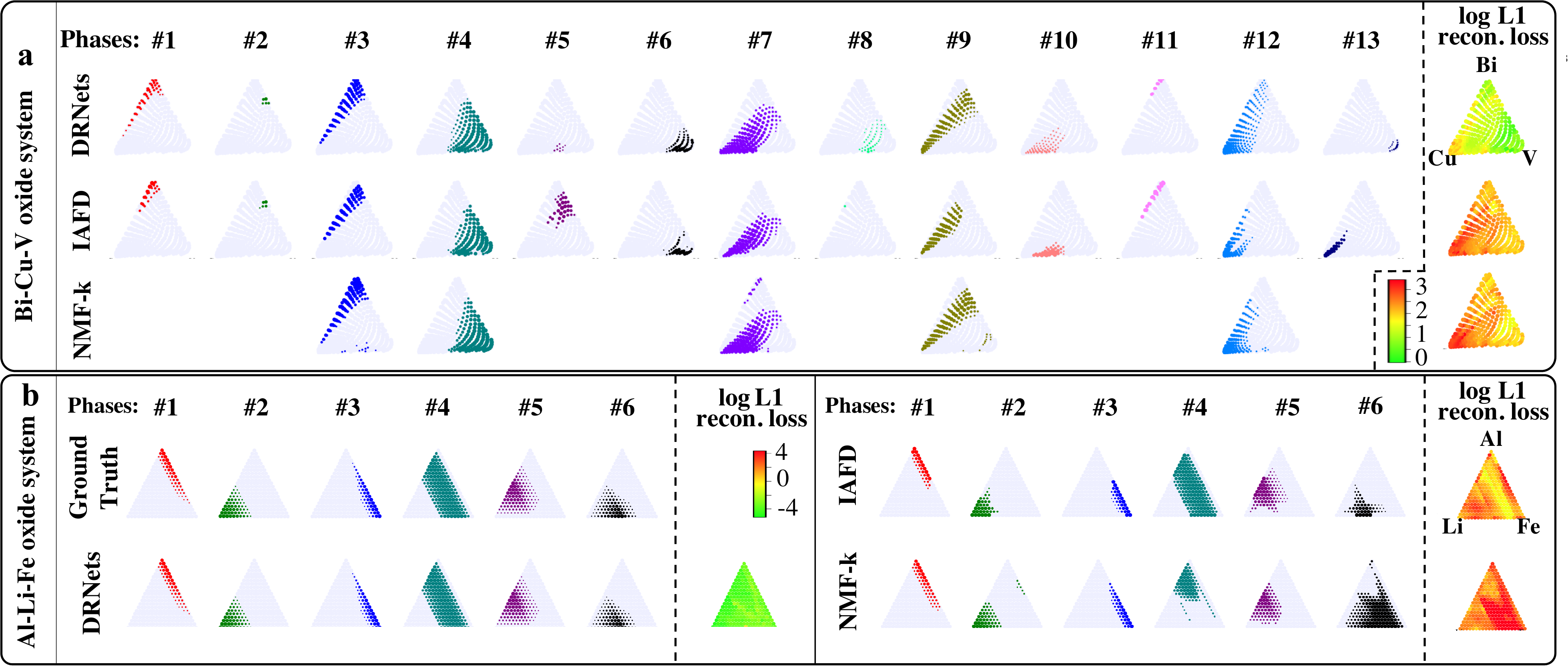}
    %\linewidth]{FIGS/pm-new-fig.png} 
  \caption{
  \textbf{
  \Carla{
Comparison of the activation maps produced by DRNets with other unsupervised approaches for the Bi-Cu-V oxide and Al-Li-Fe oxide systems.}
  } 
  \Carla{(\textbf{a}) The activation map of the 307 composition points of the  Bi-Cu-V oxide system for each of the 13 phases identified by DRNets is shown, with comparison to IAFD and NMF-k solutions (see Extended Data Fig.~\ref{fig:extra-bi-cu-v-o-drnet}), demonstrating their ability to capture some aspects of the phase activations while misrepresenting or omitting several phases that are key to generating a meaningful phase diagram. The reconstruction loss for each pattern is also shown  demonstrating that only through correct identification of the phases can the XRD dataset be fully explained. 
  In (\textbf{b}), we highlight the performance of the different methods on the  synthetically generated Al-Li-Fe oxide system (231 composition points), which has ground truth: DRNets are  the only  system that  nearly 
perfectly identifies the phases present in every XRD pattern. \Carlanew{The different methods share a common color scale for reconstruction loss in each system, and the elemental labels for the composition triangle are only provided once per system.}
See further details in 
%Extended Data Fig.~\ref{fig:extra-bi-cu-v-o-drnet} and 
Fig.~\ref{fig:down-scale-PM}a.}
  }
  \label{fig:phase-diagram-main}
\end{figure}

\begin{figure}[H]
  \centering \includegraphics
  [width=0.72\linewidth]{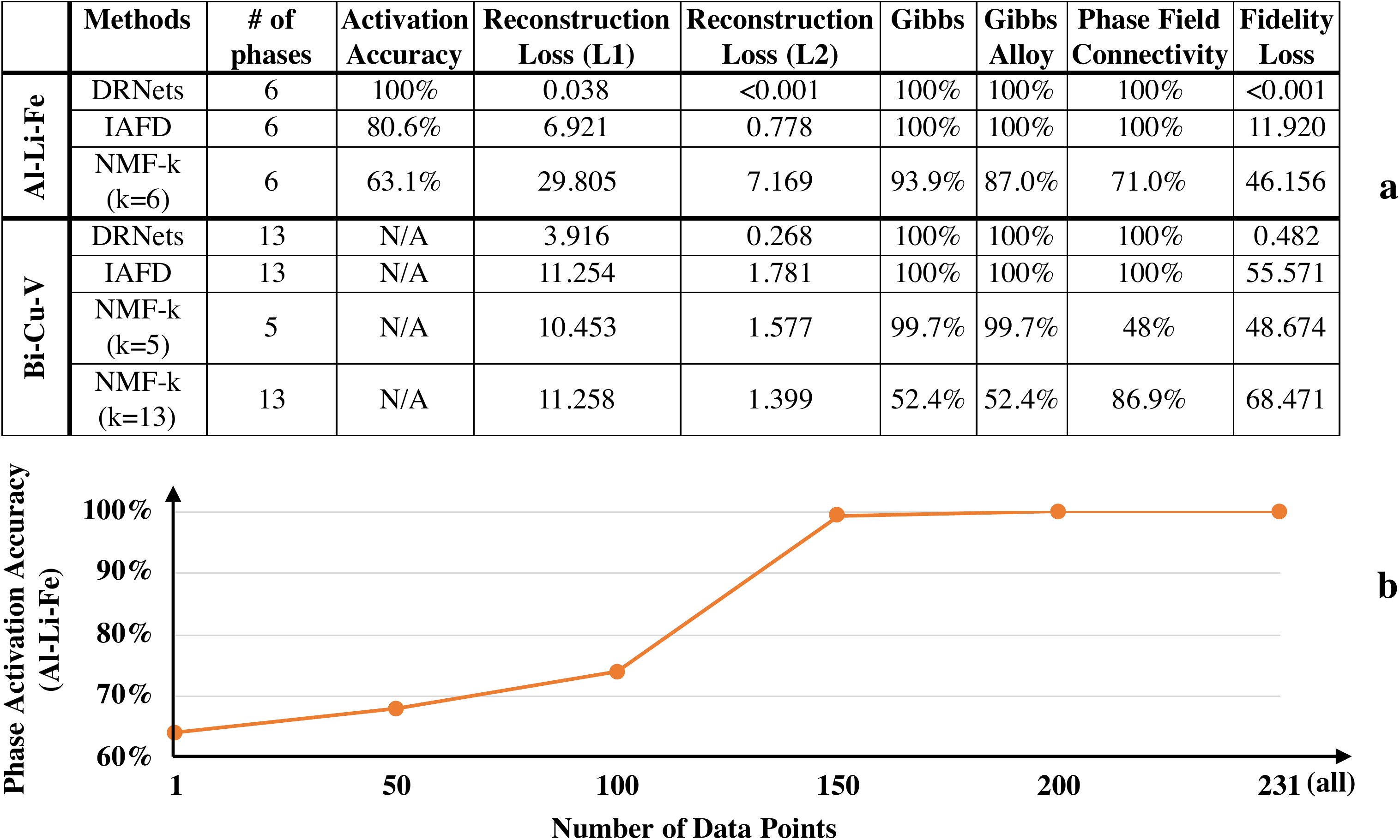} 
  \caption{
  \textbf{a. Comparison of different performance metrics for different methods for the Al-Li-Fe oxide system and the Bi-Cu-V oxide system.}
Gibbs, Gibbs alloy, and phase connectivity metrics denote the proportion of samples satisfying the Gibbs, Gibbs alloy, and phase connectivity rules; the phase fidelity loss (Fidelity Loss) denotes how well the discovered phase patterns match the ground truth (the lower the better 
 \hyperref[Method:phase_mapping]{(See Supplementary Methods)}).
%\john{(see Supplementary Methods).}
%, definition in \hyperref[SI:phase_mapping]{ SI~\ref{SI:phase_mapping}}). 
For the Al-Li-Fe oxide system, the DRNets solution has 6 phases, which matches ground truth, and this known number of phases was applied to IAFD and NMF-k.
%for all the methods, the number of pure phases is 6, which is the number of phases in the ground truth.
%
%
For the Bi-Cu-V oxide system, both DRNets' and IAFD's solutions have 13 phases (the number of phases was specified for IAFD but not DRNets) while NMF-k has 5 phases. Nevertheless, we also run NMF-k with 13 phases following the verification of the presence of the 13 phases from the DRNets solution.
%, given that materials science experts confirmed the existence of 13 phases, as outputed by DRNets.
%
Note that, there is no ground truth for the Bi-Cu-V oxide system, therefore the activation accuracy is not applicable (N/A).
The results indicate that DRNets perform substantially better than IAFD and NMF-k for all the metrics on both systems (Additional  details
in %\john{Supplementary Methods}
\hyperref[Method:phase_mapping]{Supplementary Methods}).
%\hyperref[SI:phase_mapping]{ SI~\ref{SI:phase_mapping}}).
%
\textbf{b. The performance of DRNets on Crystal-Structure Phase Mapping (Al-Li-Fe oxide system) with different number of XRD data points.}
 Learning over multiple XRD patterns within a composition system plays an important role for DRNets to solve crystal-structure phase mapping problem. 
 As shown in the plot, for Al-Li-Fe oxide system, DRNets can almost perfectly recover the phase activation of XRD patterns when  it  
learns via demixing of a collection of at least
 %it is learned over more than
 150 XRD patterns.
  }  \label{fig:down-scale-PM}
\end{figure}

%%%%
Since ground truth is unavailable for previously-unsolvable experimental phase mapping datasets, we first compare DRNets to state of the art methods NMF-k\cite{stanev2018unsupervised} and IAFD\cite{gomes2019crystal} using a synthetic benchmark dataset (with ground truth), based on the Al-Li-Fe oxide system, \cite{stanev2018unsupervised,le2014challenges} which contains 6 phases in 15 unique combinations (from  159 prototypes) with substantial alloying.
%(see 
%Extended Data Fig.~\ref{fig:Li-Fe-Al-map} 
%Fig.~\ref{fig:bi-cu-v-o}g
%and
%Extended Data Table \ref{table:PM_comparison}). 
DRNets outperform NMF-k and IAFD in a variety of metrics.
We also  considered a recently proposed supervised algorithm\cite{lee2020deep} in which a deep neural network, trained using
XRD patterns simulated from prototypes of known phases,
%simulated data based on known crystal phase patterns,
directly predicts the phases present in a given XRD pattern. While such an approach can be effective for complementing human expertise for a single XRD pattern, it performed poorly on a complex system such as the benchmark Al-Li-Fe oxide system (phase identification accuracy around 1\%),  which exposes the limitations of a purely simulation-based supervised approach for handling the combinatorics of phase mapping (details in \hyperref[Method:phase_mapping]{Supplementary Methods}).
%\john{Supplementary Methods}
%\hyperref[SI:phase_mapping]{SI~\ref{SI:phase_mapping}}
%These results are emblematic of the challenges facing incorporation of deep learning in science, the lack of sufficient training data to learn the rules of science. 
In contrast, \Carlanew{the  DRNets  approach is the only one that 
perfectly} identifies the phases present in every XRD pattern \Carlanew{and learns the phase-pure patterns (Extended Data Fig. \ref{fig:Li-Fe-Al-phases}). The DRNets model} outperforms other algorithms in a variety of other metrics (See Fig.~\ref{fig:phase-diagram-main}b and Fig.~\ref{fig:down-scale-PM}a). \Carla{In an ablation study we show that DRNets  need  around 150 input XRD patterns to approach the ground truth solution in the Al-Li-Fe oxide system (Fig.~\ref{fig:down-scale-PM}b). This  study highlights the importance of the data-driven learning of the shared parameters of the demixing task across multiple XRD patterns, 
%wherein the combinatorial constraints applied to a sufficiently large set of interrelated XRD patterns 
which enables each pattern to be demixed in a manner that is often not possible with single isolated XRD patterns. Nevertheless, DRNets' data requirements (hundreds of data points) are considerably smaller than those of standard deep learning approaches (hundreds of thousands of data points).}
%See Fig.~\ref{fig:phase-diagram-main}, and Extended Data Table \ref{table:PM_comparison} and Extended Data Fig.~\ref{fig:down-scale-PM}.

To represent unsolved experimental datasets, we use the Bi-Cu-V oxide system where manual analysis was found to be particularly ineffective \Carlanew{to solve the system} due to the complexity of the alloying in the set of 307 XRD patterns, as well as strong overlap of signals in the 100 prototypes 
%Fig.~\ref{fig:bi-cu-v-o}, Fig.~\ref{fig:phase-diagram-main}a and Extended Data 
(Fig.~\ref{fig:sudoku}a-f and Fig.~\ref{fig:extra-bi-cu-v-o-drnet}). DRNets identified 13 phases in 19 unique mixtures (Fig.~\ref{fig:phase-diagram-main}a and Fig.~\ref{fig:extra-bi-cu-v-o-drnet}), and the presence of each phase was verified by manual analysis using standard practices based on absence of missing peaks and inability to explain the signal with other prototypes. \Carlanew{We note that in practice verifying a solution is easier %the
 than 
producing it. For example, visual inspection of Fig.~\ref{fig:extra-bi-cu-v-o-drnet} reveals the excellent agreement between the stick prototype and the demixed DRNets model for each phase, and this analysis was extended to patterns from the experimental dataset that were chosen based on high activation of each phase and each phase mixture, a manual validation based on representative patterns from the solution. }
%\john{DELETE[[[Importantly, the manual pattern verification was uniquely enabled by the DRNets solution.]]]}
\Carla{To assess the extent by which learning the pattern interrelationships is critical for solving the Bi-Cu-V oxide system, we consider a model analogous to DRNets for demixing a single XRD pattern in isolation.} \Carlab{This model identifies the same phases as DRNets for only 27\% of the patterns, highlighting that the nuanced phase behavior of this system can only be resolved through combinatorial experimentation 
%matched with  learning the shared parameters across multiple XRD patterns 
combined with reasoning about the underlying  thermodynamic  constraints  to learn the shared parameters across multiple XRD patterns.}
%Importantly, the identification of patterns for manual analysis was uniquely enabled by the DRNets solution.
As shown in Fig.~\ref{fig:phase-diagram-main}a, the DRNets activation maps (the amount of each demixed pattern in each input XRD pattern) for 5-8 of the phases are \Carla{poorly} reproduced by NMF-k and IAFD. The demixed patterns in these solutions\Carlanew{, which are intended to be phase-pure patterns,} contain mixtures with the less-commonly-occurring phases, making the minor phases undiscoverable by these methods and hampering inference of scientific knowledge from the phase mapping solution. The ``fidelity loss'' quantifies the deviation of the demixed patterns from its closest prototype (see 
%\john{Methods}
\hyperref[Method:phase_mapping]{Supplementary Methods}
%\hyperref[sec:methods]{Methods}
), and the lack of phase purity in the demixed NMF-k and IAFD solutions contribute to their substantial fidelity loss compared to DRNets (Fig. \ref{fig:down-scale-PM}a). 
\Carlanew{The low fidelity loss of the DRNets solution is commensurate with the manual verification of the presence of phases identified by DRNets, although this analysis does not preclude the false negative detection of a phase in any given XRD pattern. Such an imperfection in the solution would give rise to a reconstruction error, and Fig. \ref{fig:down-scale-PM}a demonstrates that the reconstruction loss for DRNets is substantially lower than those of other methods, indicating that false negative phase detection is not a major issue in the DRNets solution. Substantial reconstruction loss can also occur if the experimental data contains a phase that is missing from the set of prototypes, which would prompt an investigation of phase discovery, as discussed further in the \hyperref[Method:phase_mapping]{Supplementary Methods}. 
The activation accuracy metric assesses the phase concentrations in each measured pattern but can be quantified only when ground truth is available, as in the synthetic dataset where DRNets substantially outperform other methods.
\Carlanew{Datasets with poor signal-to-noise ratio and/or XRD peak widths that do not enable unambiguous phase identification can result in different solutions that still satisfy thermodynamic rules and reconstruct the source data\cite{gomes2019crystal}. 
% requiring additional experimentation. %to find a unique solution, 
Evaluating the stability of DRNets' phase mapping solutions with active feedback from experiments is
%and active iteration with data acquisition comprise
an interesting avenue that we are pursuing.}
%are not a manual inspection of the individual patterns with highest reconstruction loss revealed the presence of peaks with intensities near the detection limit that are not explained by the DRNets solution. These small residual peaks are insufficient for definitive phase identification, indicating that minor imperfections in the DRNets solution may exist but are not quantifiable beyond the metrics of Fig. \ref{fig:down-scale-PM}a. The activation accuracy metric can be quantified only when ground truth is available, as in the synthetic dataset where DRNets substantially outperform other methods. 
}

Informed by the DRNets solution, analysis of the photo-oxidation of water (a critically limiting component of solar fuels technology) revealed that a 3-phase mixture (alloy variants of  BiVO$_4$, Cu$_2$BiVO$_6$ and Cu$_3$(VO$_4$)$_2$) outperforms the standard monoclinic BiVO$_4$ material, defying the conventional wisdom that phase mixtures are deleterious to photoactivity (Extended Data Fig.~\ref{fig:bi-cu-v-pec}). 
%DRNets, self-supervised by scientific prior knowledge, enable principled incorporation of deep learning to accelerate knowledge discovery.
DRNets' performance for phase mapping is emblematic of how seamlessly combining deep learning with  reasoning about prior scientific  knowledge can automate the interpretation of scientific data and accelerate knowledge discovery.

%% file: new-discussion.tex
\begin{comment}
\begin{figure}[h]
\centering
\includegraphics [width=1
\linewidth]{FIGS/summary-table-v4.png} 
\caption{\small{\textbf{Different components of DRNets for the different tasks (not sure about this table). }}
   }
\label{fig:table}
\end{figure}
\clearpage
\end{comment}
  %\textit{Seeing further   by standing on the shoulders of Giants} and interpretability are hallmarks of the scientific process. Moreover, science has been successful in describing the world and capturing natural phenomena through
 \section*{\small{Discussion}}
\Carla{A central tenet of the scientific process is the interpretation of data in the context of a rich body of scientific knowledge. However, the efficient integration of complex prior knowledge into machine learning approaches has been an open  challenge in AI. 
  DRNets  provide a general modular framework for incorporating  prior knowledge that can be customized to effectively tackle \revised{ unsupervised pattern demixing tasks.
  Herein we provide an in-depth description of the application of DRNets to  a  scientific application, crystal-structure phase mapping, a fundamental  long-standing \Carlab{challenge} in  materials science. We also illustrate the application of  DRNets to  a Sudoku demixing task, which is facilitated by the availability of  benchmark data for algorithm comparisons and ablations studies. }
  %All the DRNets applications discussed in this manuscript revolve around an \textit{interpretable} structured latent space, which is key for DRNets to be able to incorporate and reason about domain and scientific background knowledge: 
  In the DRNets framework, an \textit{interpretable} structured latent space is crucial for the incorporation of  
   background knowledge (see Fig.~\ref{fig:latent_space} and Extended data Fig.~\ref{fig:table}).
  %In all the applications considered, 
  \Carlad{\textit{The semantics of  the latent space  emerges during  the optimization process as the result of the interplay  among  the encoder,  the generative decoder,  the reasoning module, and (unlabeled)  data.}} We showed the effectiveness of our approach, 
  %in the absence of labeled training data, 
  such as for crystal-phase mapping, where DRNets  significantly outperformed previous approaches and solved previously unsolved complex chemical systems, aiding
%enabled 
the discovery of solar-fuels materials. 
  %, aiding the discovery of solar-fuels materials.
  %Nevertheless, we note that the intended  latent space semantics  is not guaranteed to happen,
  %as the optimization can get stuck in local minima, 
  %and so it is even somewhat surprising how well it works for all the domain applications considered herein, especially for the Sudoku and crystal-phase mapping applications, with no training data, in which cases DRNets significantly outperformed previous approaches and solved previously unsolved complex chemical systems.
  Intuitively, the strong prior knowledge injected via the \Carlanew{fixed} generative decoder and reasoning constraints, combined with the data, %provide efficient guidance 
  constrain the optimization process to enforce the intended semantics of the latent space.
 As our Sudoku ablation  studies show, if we either replace the \Carlanew{fixed} cGAN with a (weaker) standard  learnable decoder, without prior knowledge about single digits, or remove the reasoning modules, the optimization process can no longer find the right semantics for the latent space, and the digit and Sudoku accuracy deteriorate significantly.  Data are critical, as they ultimately determine the semantics of the latent space, and  DRNets  clearly benefit from learning across  multiple (unlabeled) instances. Nevertheless, somewhat surprisingly, the amount of unlabeled data required in DRNets  for   both the Sudoku and crystal-structure phase mapping applications is considerably more modest %\Carlab{(in the orders of a few hundred samples)}
 than in standard supervised deep learning settings. 
 %Nevertheless, somewhat surprisingly, DRNets can achieve high accuracy  with only a few (unlabeled) instances,  which is in contrast with the standard deep learning approaches, which  typically require much larger datasets. 
 %There are many research directions for extending the DRNets framework, as it is general and modular and therefore it should be possible to  customize it to other  demixing  applications  and other  unsupervised tasks for which domain rules and pattern prototypes or generative models are available.
%DRNets are  a modular   end-to-end framework  that can be customized for a range of  pattern demixing  tasks in various application domains for which rich prior knowledge is available.
%, whic 
%as domain rules and pattern prototypes or generative models, which are 
%for   the combination of pattern recognition and
% prior
%knowledge
%reasoning capabilities, 
%pervasive in various application domains.
}
 
\Carlanew{
%There are many research directions for extending the 
The DRNets framework is general and modular 
and can be adapted
to other  de-mixing  and   unsupervised tasks for which domain rules and pattern prototypes or generative models are available. 
%Extending DRNets for more challenging  reasoning tasks with more powerful search methods such as deep reinforcement learning
 %to further exploit prior knowledge
% is another research direction. 
In addition,  
%while  prior knowledge is  important to compensate for the lack of training data  and condition the models to produce physically meaningful solutions in unsupervised settings,  
prior knowledge can also  increase interpretability and boost performance of  supervised models, another research directions for DRNets.
%, by  restricting the parameter space to more realistic settings, which often is not fully captured by the labeled data. 
In fact, some of the ideas in DRNets were 
%actually
inspired by our  work in the context of supervised learning on multi-label classification and in particular deep-learning-based joint species distribution models (JSDMs)\cite{chen2016deep} \Carlanew{that integrate }
%integrating
prior ecological knowledge and species observational training data  from the eBird citizen science program.~\cite{sullivan2014ebird}  %
%Communities of species are structured by 
The need to capture and interpret
interactions between species and their local environments as well as interactions among different species,  %The ability to discern between these processes is essential for addressing many of the 
 which are  core questions in ecology and conservation,~\cite{sutherland2018horizon}
%(shared covariance matrix) 
was the initial motivation and inspiration for the semantically meaningful structured latent-space
in
%for 
the DRNets framework.\cite{chen2016deep} }
\Carlanew{We anticipate that these concepts will extend to} \Carlanew{other tasks, for example,  materials science tasks beyond phase mapping, such as property prediction. The construction of a latent space with intended semantics can not only enable constraint reasoning, as demonstrated in the present work, but also facilitate transfer learning wherein relationships among composition, structure, and some types of properties can be used to learn relationships for other types of properties. Such concepts extend the purview of prior knowledge from the rules and prototypes employed in the present work to rules, surrogate models, etc. from ancillary domains.}
\Carlanew{%Extending the  DRNets framework for supervised learning by  seamlessly integrating 
%logical and constraint reasoning and more generally 
%physics-based prior knowledge into deep learning is  an important research direction. %in particular when only small labeled training datasets are available and there is considerably  prior knowledge that can be exploited to further condition and  improve interpretability and performance.
More generally, research 
on incorporating %in a principled way 
neural network-based learning with symbolic knowledge representation and logical reasoning
%a rich body of prior knowledge, data, reasoning, optimization, learning, 
%on neurosymbolic combining   techniques to tackle  applications in  scientific domains, 
is an important next frontier in AI/ML research.~\cite{garcez2020neurosymbolic}  The  DRNets framework represents a step in that direction, providing  a modular   end-to-end framework  that can be customized for a range of  tasks that require the combination of learning and
% prior
%knowledge
reasoning, which are pervasive across a variety of application domains.}

%% file: methods.tex
\newpage
\begin{methods}
\label{sec:methods}
%NMI "Articles can also contain a Methods section, which should appear after the main text and should typically not exceed 3,000 words."
%\subsection{Mathematical formulation of DRNets.}
\label{subsec:formulation}
%Deep Rasoning Nets (DRNets) provide a rich framework integrating data-driven deep learning with logic and constraint reasoning for incorporating prior knowledge. 
%At a high level, DRNets combine deep learning with  constraint 
%reasoning. 
%\john{Herein we describe the mathematical formulation and implementation of DRNets, with further description of the applications in Supporting Methods}. 
\textbf{\Carlanew{Mathematical formulation of DRNets.}} In order to produce a DRNets encoding for a given task, we start by formulating the task as a data-driven constrained optimization problem, which is then 
transformed
%reduced
through a sequence of steps into a data-driven unconstrained optimization problem amenable to end-to-end optimization via state-of-the-art deep learning technology 
(Extended Data Fig.~\ref{fig:DRNet-flow}a). \revised{More formally,
%the DRNets' formulation of a task as data-driven constrained optimization problems is as follows:
the formulation of a DRNets task as 
a data-driven constrained optimization problems is as follows:}
\begin{comment}
Deep Rasoning Nets provide a rich framework integrating deep learning
with logic and reasoning for incorporating prior knowledge
constraints.  More formally, to encode a problem in the DRNets framework, we start by formulating  tasks as
constrained optimization problems, incorporating abstractions and
reasoning about structure and prior knowledge. See Extended Data Fig.~\ref{fig:DRNet-flow}. 
\end{comment}
\revised{
\small
%\begin{linenomath}
\begin{gather}
    \min\limits_{\theta}\; \frac{1}{N}\sum^N_{i=1}{\mathcal{L}(G(\phi_\theta(\mathbf{x}_i)), \mathbf{t}_i)} \quad
    %\mbox{subject to } \{\phi_\theta(\mathbf{x}_i)\}^N_{i=1} \in \Omega
    \mbox{ subject to: } \phi_\theta(\mathbf{x}_i) \in \Omega^{\mbox{local}} \mbox{ and }
    (\phi_\theta(\mathbf{x}_1), ..., \phi_\theta(\mathbf{x}_N) ) \in \Omega^{\mbox{global}}
     \nonumber\\
     \mbox{where }\phi_\theta(\mathbf{x}_i)\defeq(\mathbf{z}_{i,1},...,\mathbf{z}_{i,m}, \mathbf{e}_{i,1},...,\mathbf{e}_{i,m})
    \label{eqn:constrainted}
\end{gather}
%\end{linenomath}
\normalsize
In this formulation, }
\revised{
$N$ is the number of input data points, 
$m$ is the number of possible single patterns,}
$\mathbf{x}_i \in R^n $ is the $i$-th
$n$-dimensional input data point, \revised{$\mathbf{t}_i$ is the corresponding targeted output, which is in general the input $\mathbf{x}_i$ in unsupervised cases,
$\phi_\theta(\mathbf{x}_i)\defeq(\mathbf{z}_{i,1},...,\mathbf{z}_{i,m}, \mathbf{e}_{i,1},...,\mathbf{e}_{i,m})$ is the latent space of DRNets for data point $\mathbf{x}_i$, a function
of the encoder $\phi$ parameterized by $\theta$, typically a neural network.
$\mathbf{z}_{i,j}$ and $\mathbf{e}_{i,j}$ are the \textit{shape} and  \textit{probability embeddings} of the possible pattern-$j$ at data point $\mathbf{x}_i$ indicating its %possible 
shape and probability.
$G(\cdot)$ is the generative decoder which involves a fixed pre-trained or parametric generative model that generates single patterns from shape embeddings $\mathbf{z}_{i,j}$
and a process that mixes the  generated single patterns factoring in their probabilities.
% $\phi_\theta(\cdot)$ is the \revised{latent space of DRNets, a function
% of the encoder parameterized by $\theta$,} typically a neural
% network, $G(\cdot)$ denotes the generative decoder,
$\mathcal{L}(\cdot,\cdot)$ is the loss function, \revised{which evaluates the 
loss between the output of the generative decoder and the target $\mathbf{t}_i$,}
%reconstruction of patterns
 $\Omega^{\mbox{local}}$ and
$\Omega^{\mbox{global}}$ are the constrained spaces w.r.t.\ a single
input data point and several input data points, respectively.
}
Note that constraints can involve several (potentially all) data points:
e.g., in Sudoku, all digits should form a valid Sudoku and in
crystal-structure-phase-mapping, all data points in a composition
graph should form a valid phase diagram. Thus, we specify local and
global constraints in DRNets -- local constraints only involve a
single input data point whereas global constraints involve several
input data points, and they are optimized using different strategies.
For the Multi-MNIST-Sudoku DRNets, $\phi_\theta(\cdot)$ is composed of two ResNet-18,
\revised{The generative model in} $G(\cdot)$ is a pre-trained conditional
GAN \cite{mirza2014conditional} using hand-written digits; for the Crystal-Structure-Phase-Mapping DRNets, $\phi_\theta(\cdot)$ is composed of four 3-layer-fully-connected networks
and $G(\cdot)$ involves a Gaussian Mixture model.
Below we provide details for each specific
application. 

\revised{Finally, we note that we refer to \textit{data-driven} constrained/unconstrained optimization problems to highlight the fact that  even though we assign an interpretation to the structured latent-space produced by the encoder, $\phi_\theta(\mathbf{x}_i)$,  the  latent-space semantics are  ultimately determined by the data. }\revised{(See also Extended Data Fig.~\ref{fig:table} for a summary of the different components of DRNets for the different tasks.)}
% and applications and \hyperref[SI:InformalSudoku]{ SI~\ref{SI:InformalSudoku}} for an informal description of the \revised{mechanics of formulating DRNets for Sudoku)

\textbf{\revised{Transformation} flow for DRNets.} Solving the constrained optimization problem in equation (\ref{eqn:constrainted})
directly is challenging since the objective function in
general involves deep neural networks, which are highly non-linear and
non-convex, and prior knowledge often involves combinatorial
constraints, \revised{which cannot be directly encoded in a standard deep learning framework.}  Extended Data Fig.~\ref{fig:DRNet-flow}a depicts how we
the reduce the above constraint optimization problem into DRNets. 
In particular, we use Lagrangean relaxation to approximate the
constrained optimization problem (equation \ref{eqn:constrainted})
with an unconstrained optimization problem, moving the constraints to
the objective function with associate penalty weights:
\revised{
\small
%\begin{linenomath}
\begin{gather}
    \min\limits_{\theta}\; \frac{1}{N}%\bigg(
    \sum^N_{i=1}{\mathcal{L}(G(\phi_\theta(\mathbf{x}_i)), \mathbf{t}_i)} + \lambda^l\psi^{l}(\phi_\theta(\mathbf{x}_i))%\bigg)
    + \sum^{N_g}_{j=1}\lambda^g_j\psi^g_j(\{\phi_\theta(\mathbf{x}_k)|k\in S_j\})
    \nonumber\\
     \mbox{where }\phi_\theta(\mathbf{x}_i)\defeq(\mathbf{z}_{i,1},...,\mathbf{z}_{i,m}, \mathbf{e}_{i,1},...,\mathbf{e}_{i,m})
    \label{eqn:object-0}
\end{gather}
%\end{linenomath}
\normalsize 
}
\revised{Herein}, 
%$N$ is the number of input data points, 
$N_g$
denotes the number of global constraints, $S_j$ denotes the set of
indices w.r.t.\ the data points involved in the $j$-th global
constraint, and $\psi^l, \psi^g_j$ denote the penalty functions for
local constraints and global constraints, respectively, along with
their corresponding penalty weights $\lambda^l$ and $\lambda^g_j$.  As
outlined in  Extended Data Fig.~\ref{fig:DRNet-flow}a, we employ
two mechanisms to tackle the above unconstrained optimization task:
(1) \textit{continuous relaxations} of constraints with discrete variables and (2) \textit{constraint-aware stochastic gradient descent} to tackle global penalty functions, \revised{for the different types of combinatorial constraints.} 

 \label{methods:relax}
\textbf{Continuous Relaxations:} Prior knowledge often involves
combinatorial constraints with discrete variables that are difficult
to optimize in an end-to-end manner using gradient-based methods.
Therefore, we need to design proper continuous relaxations for
discrete constraints to make the overall objective function
differentiable. 
Many approaches have been used to handle continuous relaxations of discrete constraints \cite{hu2016harnessing,xu2017semantic}.
%,wilder2018melding,wang2019satnet}
%\citet{hu2016harnessing} proposed a continuous relaxation for first-order logic and applied it on a sentiment analysis task.
%For tasks in our paper, 
We apply a group of entropy-based continuous relaxations to encode
general discrete constraints such as sparsity, cardinality, and 
All-Different constraints.
%, and SAT constraints (see
%Fig.\ref{fig:relaxation}). 
%Moreover, our framework can be easily
%expanded to encode other constraints or other relaxations. %for e.g., \textit{linearprograming}, \textit{mixed integer programming}, and \textit{quadratic     programming}.
%practitioners can follow our examples to design their own relaxations for incorporating other types of constraints for their applications.
%Limited by space, we only discuss those relaxations in this paper
%, which states that a set of labels have to be all different.
We construct continuous relaxations based on probabilistic modelling
of discrete variables, where we model a probability distribution over
all possible values for each discrete variable. 
  For example, in
Multi-MNIST-Sudoku, a way of encoding the possible two digits in the
cell indicated by data point $x_i$ (one from $\{1...4\}$ and the other
from $\{5...8\}$),
%which is the pattern from the cell of the overlapped Sudoku, 
is to use 8 binary variables $e_{i,j}\in\{0, 1\},$
%to indicate the existence of 8 possible digits 
while requiring $\sum^4_{j=1}e_{i,j} = 1$ and $\sum^8_{j=5}e_{i,j} = 1$.
%In contrast,
In DRNets, we model probability distribution $P_i$ and $Q_i$ over
digits 1 to 4 and 5 to 8 respectively: $P_{i,j}
,\scriptstyle{j=1...4}$ and $Q_{i,j}, \scriptstyle {j=1...4}$ denote
the probability of digit $j$ and the probability of digit $j+4$,
respectively.  We approximate the cardinality constraint of $e_{i,j}$
by minimizing the entropy of $P_i$ and $Q_i$, which encourages $P_i$
and $Q_i$ to collapse to one value. See details concerning relaxations
for other constraints in Extended Data Fig.\ref{fig:DRNet-flow}b and 
%\john{Supplementary Methods.}
%\hyperref[SI:relaxation]{ SI~\ref{SI:relaxation}}. 
\hyperref[sec:si]{Supplementary Methods}. 

% {\color{blue} ChenDi, as we discussed, the section on
% continuos relaxation in the supplementary materials (SI) needs to be
% extended and better structured, including adding the table withe the
% different relaxations. In any case we need to stress that other
% relaxatins can be used. Also, I dumped the thermodynamic rules in the
% relaxation part but perhaps it's better to have more detials about
% each application and therefore the thermodynamic rules should be moved
% there.}

\textbf{Constraint-Aware Stochastic Gradient Descent:}
\label{SI:SGD}
\Carlanew{We developed a variant of the standard SGD method that we refer to as constraint-aware SGD, which batches data points involved in the same global constraint together, conceptually similar to the optimization process in GraphRNN \cite{you2018graphrnn}, to tackle
the optimization of global penalty functions
$\psi^g_j(\{\phi_\theta(\mathbf{x}_k)|k\in S_j\})$, which involve several (potentially all) data points. 
% More detailed algorithmic description in \hyperref[SI:SGD]{SI~\ref{SI:SGD}}.
\begin{algorithm}[ht]
\caption{Constraint-aware stochastic gradient descent optimization of deep reasoning networks.
%For ease of presentation, we demonstrate the case with only one local constraint and one global constraint.
%\textbf{Prerequisites:} 1. Constructed penalty functions $\psi^l(\cdot)$ and $\psi^g(\cdot)$ for the local constraint and the global constraint.
%2. Constructed "if-then" conditions based on prior knowledge.
}
\label{alg:SI_training} 
\begin{algorithmic}[1]
\REQUIRE \textbf{(i)} Data points $\{x_i\}_{i=1}^N$. \textbf{(ii)}  Constraint graph. \textbf{(iii)} Penalty functions $\psi^l(\cdot)$ and $\psi_j^g(\cdot)$ for the local and the global constraints.
\textbf{(iv)} Pre-trained or parametric generative decoder $G(\cdot)$.
%\textbf{(iv)} The constraint graph.
%\STATE Construct the penalty function $\psi^l(\cdot)$ and $\psi^g(\cdot)$ for the local constraint and the global constraint.
%\STATE Construct the "if-then" conditions based on prior knowledge.
\STATE Initialize the penalty weights $\lambda^l, \lambda_j^g$ and thresholds 
for all constraints.
\FOR{number of  optimization iterations}
\STATE Batch data points $\{\mathbf{x}_1,...,\mathbf{x}_m\}$ from the sampled (maximal) connected components.
\STATE Collect the global penalty functions $\{\psi^g_j(\cdot)\}^M_{j=1}$ concerning those data points.
\STATE Compute the latent space $\{\phi_\theta(\mathbf{x}_1), ...,\phi_\theta(\mathbf{x}_m)\}$ from the encoder.
%\STATE Check "if-then" logics and adjust the penalty weights $\lambda_l, \lambda_j^g$ and thresholds accordingly.
\STATE Adjust the penalty weights $\lambda_l, \lambda_j^g$ and thresholds 
accordingly.
\STATE minimize $\frac{1}{m}\big(\sum^m_{i=1}{\mathcal{L}(G(\phi_\theta(\mathbf{x}_i)), \mathbf{x}_i)} + \lambda_l\psi^{l}(\phi_\theta(\mathbf{x}_i))\big) +
    \sum^{M}_{j=1}\lambda^g_j\psi^g_j(\{\phi_\theta(\mathbf{x}_k)|k\in S_j\})$
    %+ \lambda_g\psi^{g}(\phi_\theta(\mathbf{x}_1), ..., \phi_\theta(\mathbf{x}_m))$ 
    using any standard gradient-based optimization method and update the parameters $\theta$. 
    %We used Adam in our experiments.
\ENDFOR
\end{algorithmic}
\end{algorithm}
%\vspace{-12pt}
%\subsection*{Constraint-Aware Stochastic Gradient Descent}
}
\Carlanew{
We define a \textit{constraint graph}, an undirected graph in which each data point forms a vertex and two data points are linked if they are in the same global constraint.
%Therefore, the data points involved in the same constraint are pairwisely connected and . 
Constraint-aware SGD batches data points from the randomly sampled (maximal) connected components in the \textit{constraint graph}, and optimizes the objective function w.r.t.\ the subset of global constraints concerning those data points and the associated local constraints. 
For example, in Multi-MNIST-Sudoku, the data points (cells) in each overlapping Sudoku form a maximal connected component in the \textit{constraint graph}, so we batch the data points from several randomly sampled overlapping Sudokus and optimize the All-Different constraints (global) as well as the cardinality constraints (local) within them.
However, in Crystal-Structure Phase Mapping, the maximal connected component becomes too large to batch together, due to the constraints (\textit{phase field connectivity} and \textit{Gibbs-alloying rule}) concerning all data points in the composition graph. 
Thus, we instead only batch a subset (still a connected component) of the maximal connected component -- e.g., a path in the composition graph, and optimize the objective function that only concerns constraints within the subset (along the path).
By iteratively solving sampled local structures of the "large" maximal component, we cost-efficiently approximate the entire global constraint.
For efficiency,
%and potential capability of generalization,
DRNets solve all instances together using constraint-aware SGD (see Algorithm \ref{alg:SI_training})
}

\Carlanew{
Moreover, for optimizing the overall objective, constraint-aware SGD dynamically adjusts the thresholds and the weights of constraints according to their satisfiability, which can involve non-differentiable functions.
Specifically, we initialize the penalty weights and thresholds of constraints 
for penalty functions using hyper-parameters.
}
\Carlanew{During optimization DRNets keep track of  the satisfiability of constraints (this step could involve non-differentiable functions) and adjust the penalty weights and thresholds based on their satisfiability.}
% This mechanism is mainly designed for Crystal-Structure Phase Mapping because DRNets can already achieve perfect performance for Multi-MNIST-Sudoku with fixed weights. 
\Carlanew{For example, in crystal-structure phase mapping, we use the k-sparsity constraint to implement the Gibbs rule (the number of phases than can coexist is at most 3 in a ternary chemical system) and the Gibbs-Alloy rule (the max phase count decreases by 1 if ``alloying'' occurs) in DRNets. 
The threshold $c$ of k-sparsity is initialized as $\log 3$ (Gibbs rule), which is the entropy when the probability mass is evenly distributed among $3$ phases.
During the optimization of DRNets, if ``alloying'' occurs (by checking the phase shifting ratio between adjacent data points),} \Carlanew{DRNets decrease the max phase count from 3 to 2 and adjust the threshold $c$ from $\log3$ to $\log2$ (Gibbs-Alloy rule) accordingly.
Moreover, note that even if we optimize the penalty function ($H(P_M)$) to be equal or smaller than $\log k$ (k is 2 or 3), it could be the case that there are more than $k$ phases with positive probability mass ($>0.01$) and their probability mass may not be evenly distributed, which means the k-sparsity is still not satisfied.
Thus, DRNets keep tracking the satisfiability of k-sparsity constraint: if the entropy is already below the current threshold (e.g., $\log k$) and there are still more than $k$ phases with probability mass more than $\epsilon$ (0.01), we decrease the threshold $c$ to keep enforcing the model to minimize the entropy until reaching the k-sparsity.
}

\textbf{\Carlanew{Computational details: }} All the experiments are performed on one NVIDIA Tesla V100 GPU with 16GB memory.
For the training process of our DRNets, we select a learning rate in $\{0.0001, 0.0005, 0.001\}$ with Adam optimizer \cite{kingma2014adam}, for all the experiments.
For baseline models, we followed their original configurations and further fine-tuned their hyper-parameters to saturate their performance on our tasks.

\end{methods}

%% file: extended-data.tex
%%
%% TABLES
%%
%% If there are any tables, put them here.
%%
\clearpage
\section*{Extended Data}

\renewcommand{\figurename}{Extended Data Figure}
\label{sec:ed}
\setcounter{figure}{0}

\begin{figure}[ht]
  \centering
  \includegraphics [width=0.8
    \linewidth]{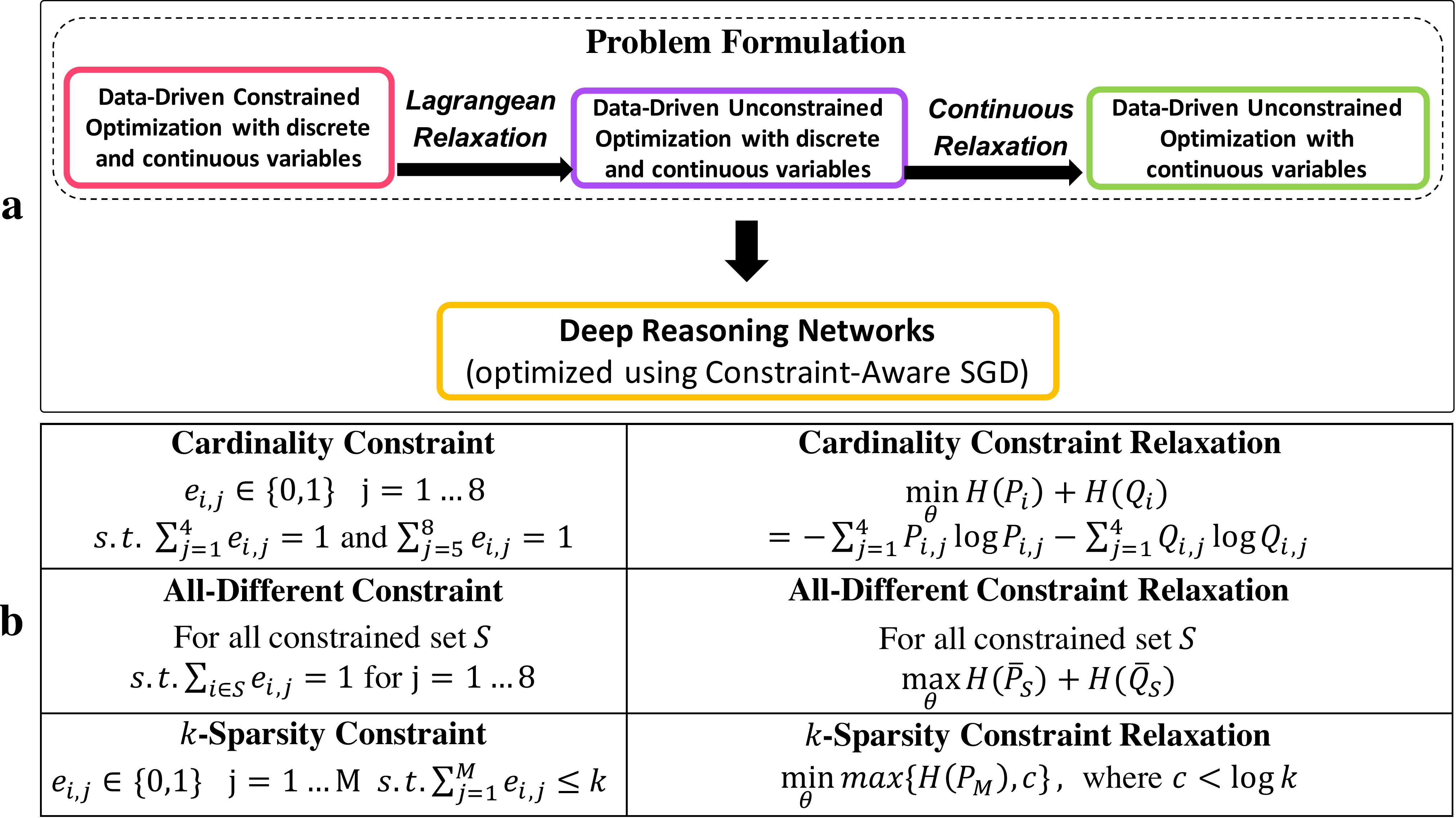} 
  \caption{ \textbf{The transformation flow for Deep Reasoning Networks and examples of continuous relaxations. %\revised{How to formulate a  DRNets application}. 
  }
  \textbf{a.} The process of formulating a problem into the DRNets framework. (i) %DRNets formulate
  We start by formulating tasks as data-driven constrained optimization problems, with discrete and continuous variables. For example, the data-driven constrained optimization task in MNIST-Sudoku is to demix images of two overlapping Sudokus such that the demixed Sudokus satisfy the Sudoku constraints and their reconstruction loss  is minimized. Furthermore, in this formulation for MNIST-Sudoku, we assume that  a  the generative decoder (cGAN) reconstructs the demixed Sudoku digit images  
  using  a  two-part latent space that encodes the digit probabilities and shapes and that the two-part latent space is  produced by two convolutional neural networks (ResNet) and is subject to the Sudoku constraints. 
  (ii) This data-driven constrained optimization problem is converted into a data-driven unconstrained optimization problem using Lagrangean relaxation, which essentially moves the constraints to the objective function, with associated penalty weights.  (iii) We use  entropy-based continuous relaxations to encode and replace  discrete (non-differentiable) constraints with continuous functions, such as sparsity, cardinality, the all-different constraint, and logical constraints. 
  %DRNets
  The objective function combines two components:  the reconstruction loss of the  generative decoder (which for Sudoku corresponds to the  reconstruction of  the demixed overlapping digit images), and the reasoning loss (which for Sudoku corresponds to the penalty weighted entropy-based  continuous function that capture the Sudoku rules). The  result of these transformations is the  DRNets  data-driven unconstrained optimization formulation.
(iv)  DRNets optimize the overall objective function using constraint-aware stochastic gradient descent (SGD). Note that we refer to  these problems as  \textit{data-driven} problems since although we assign semantics to the structured latent-space (probabilities and shapes),  their full meaning is ultimately determined by the data.
 \textbf{b.} Examples of the continuous relaxation of discrete constraints. 
 $e_{i,j}, P_i, Q_i, P_M$, and $H$,  
  %$\lambda_h$,
  %N_c, N_l,K_j,
  %$B_i$ 
  represent indicator variables denoting if a given input  image contains a given digit, the discrete distribution over digits 1 to 4, the discrete distribution over digits 5 to 8, the discrete distribution over values 1 to $M$, and the entropy function, respectively.
  %the number of clauses, the number of literals, the number of literals in the $j$-th clause, 
  %and the weights of entropy terms, 
  %respectively.
  %and the Bernoulli distribution for the $i$-th literal.  
 %   "leaky\_relu" is the leaky ReLU and let $\bar{x}_i$ be the probability that $x_i$ is false. 
 See notation and further details in
 \hyperref[sec:si]{Supplementary Methods}. %\john{Supplementary Methods.}
 %\hyperref[sec:methods]{Methods} and \hyperref[SI:relaxation]{ %SI~\ref{SI:Relaxations-Sudoku}}.
 %SI~\ref{SI:relaxation}}.
  }
    \label{fig:DRNet-flow}
\end{figure}

\begin{figure}[ht]
  \centering \includegraphics
  [width=1\linewidth]{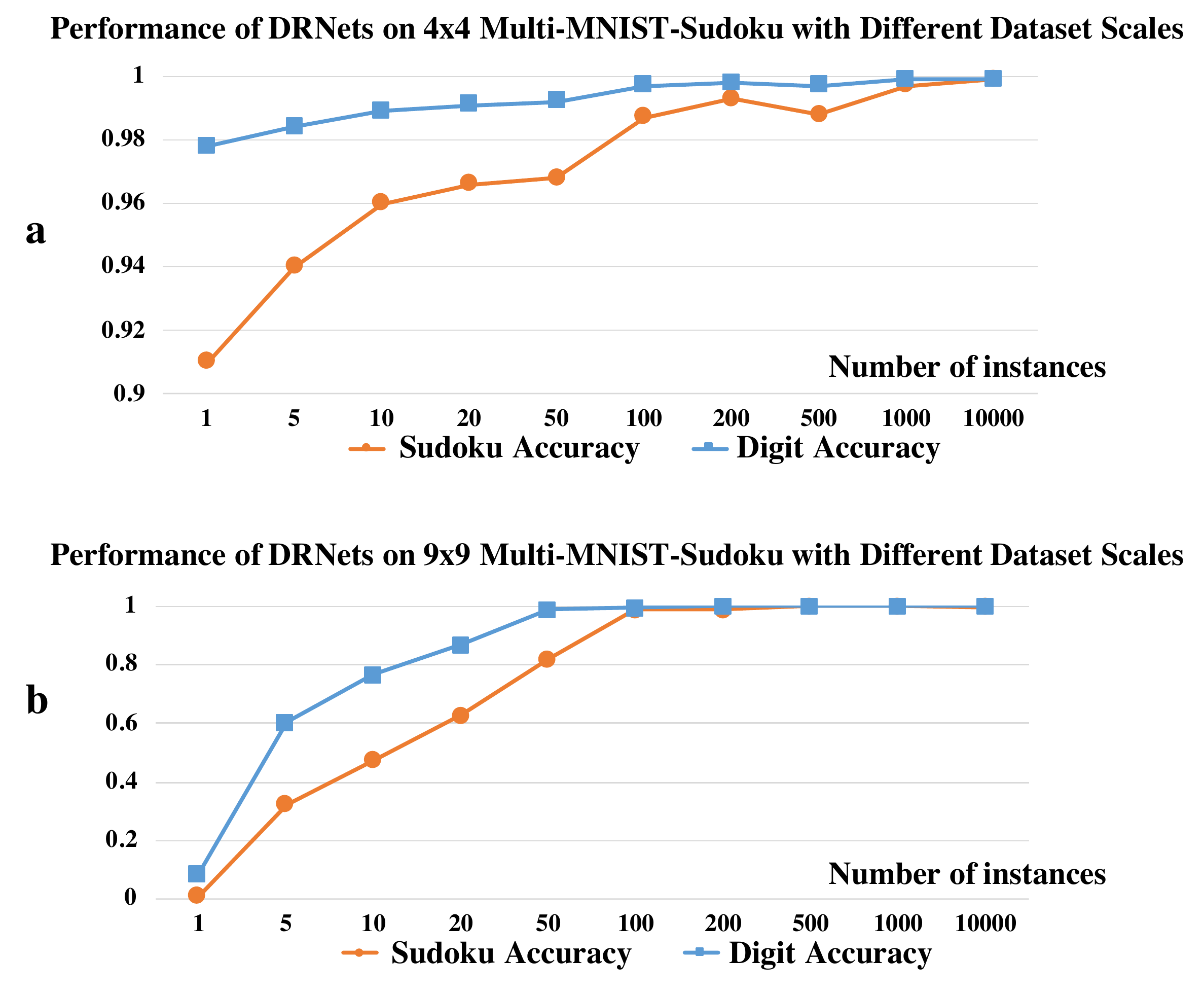} \caption{
  \revised{
  \textbf{The performance of DRNets on Multi-MNIST-Sudoku tasks with different dataset scales.}
 Learning over multiple instances significantly (especially, for the 9x9 cases) improves the performance of DRNets.
 Nevertheless, DRNets can reach 99\% Sudoku accuracy with only 100 Multi-MNIST-Sudoku instances, a considerable smaller amount of data compared to standard deep learning approaches.
  }
  }  \label{fig:down-scale-mnist}
\end{figure}

\begin{figure}[ht]
    \centering
    \includegraphics[width=1\linewidth]{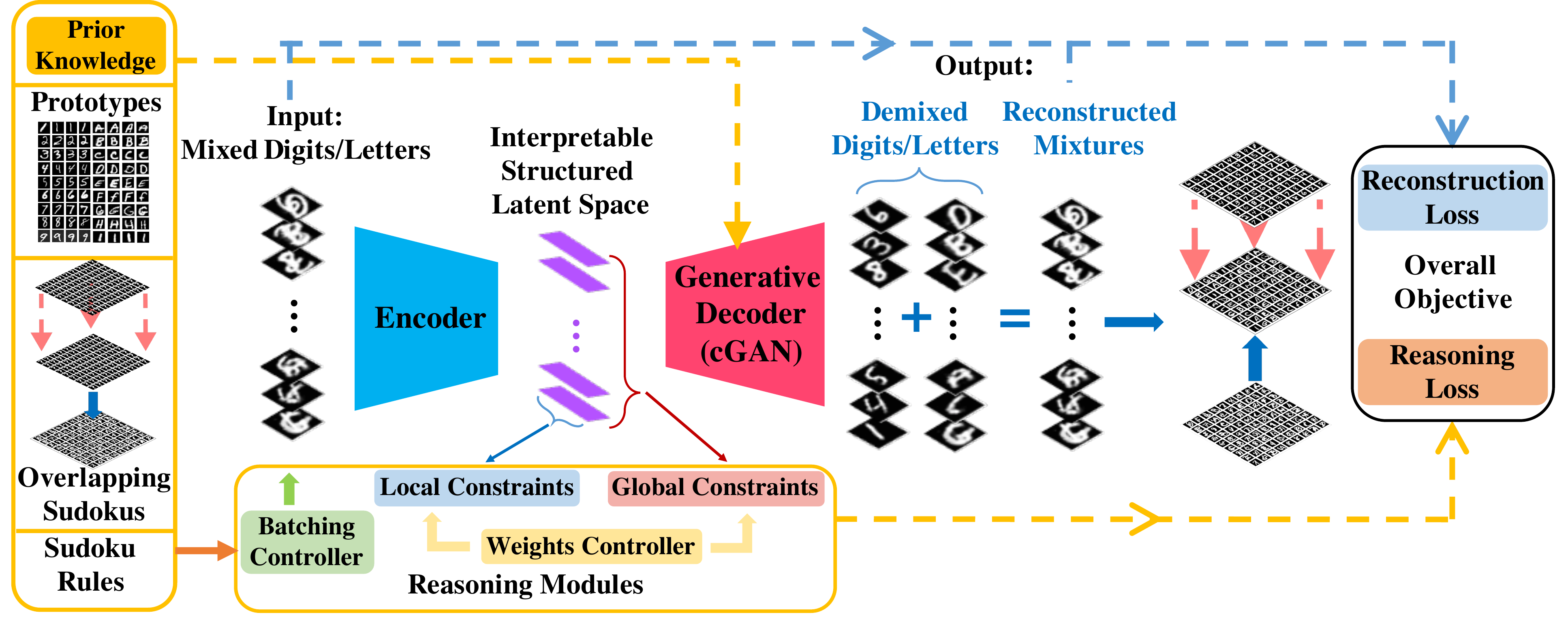}
    \caption{ 
    \textbf{DRNets for Multi-MNIST-Sudoku} DRNets perform end-to-end deep reasoning by
      using a convolutional neural network to encode an \textit{interpretable} structured latent space
      that is used by a \Carlanew{fixed} generative decoder, a conditional generative
      adversarial network (cGAN), to reconstruct the input mixed
      digits. 
      The interpretable structured latent space also allows the encoding of reasoning constraints, which enforce that the latent space adheres to prior knowledge about Sudoku
      rules.   Prior knowledge also includes digit prototypes, which
      are used to pre-train and build the \Carlanew{fixed} decoder’s generative
      module. An overall objective combines responses from the
     \Carlanew{fixed}  generative decoder 
      %\Di{to reconstruct the patterns,} 
      and the
      reasoning module and 
      %\Di{to enforce the Sudoku constraints}, and
      is optimized using constraint-aware stochastic gradient descent
      and backpropagation.
   }
   \label{fig:drnets_mnist}
\end{figure}

\begin{figure}[ht]
  \centering \includegraphics
  [width=1\linewidth]{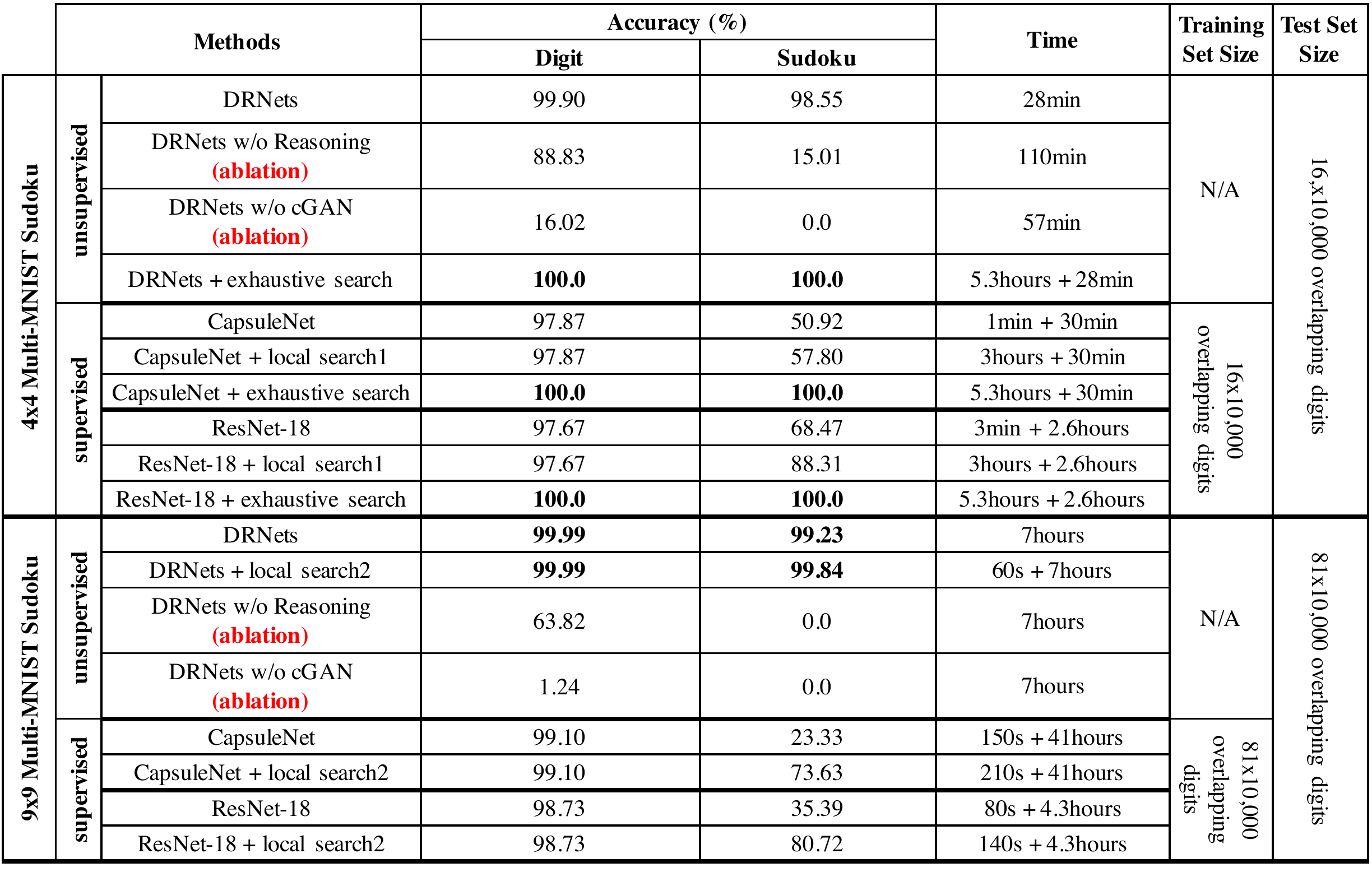} \caption{
  \small{\textbf{Comparison of the performance of different
        methods for Multi-MNIST-Sudoku 
        %(Test set: 10,000 Sudoku instances, i.e., 16$\mathbf{\times}$10,000 overlapping digits)
        }}
    % We display the DRNets' performance for two modes: optimization,
    % whose goal is to optimize for the solution, and generalization,
    % whose goal is to optimize for generalization. 
    %We compared 
    We show the
    ``solving time'' for unsupervised DRNets and its ablation variants and ``test time +
    training time'' for supervised baselines.
    % The test time for DRNets includes the optimization steps in the validation and test phase, shown in
    % parantheses. 
    The test time for CapsuleNet/ResNet + local search includes the local search time. 
    Note that we used two different local search algorithms for 4x4 cases and 9x9 cases. %"local\_search1" and "local\_search2". 
    "local\_search1" performs an enumeration for the top-2 likely digits in all 16 cells to try to satisfy Sudoku rules. 
    For 9x9 cases, it is impossible to enumerate the top-2 likely digits for 81 cells ($2^{81}$).
    Therefore, "local\_search2" conducts a depth-first search for digits in each cell from most likely to less likely until it finds a valid Sudoku combination, which is faster than "local\_search1".
    For 4x4 cases, we also applied exhaustive search for all methods, where we enumerate all possible 4x4 Sudokus and return the one with the highest likelihood given our predictions.
    Note that such strategy is not feasible for  9x9 Sudokus, given there are around $6.67\times10^{21}$ 9x9 Sudokus. The ablation study of removing the reasoning modules (DRNets w/o Reasoning)  shows that not only does the Sudoku accuracy degrades, the digit accuracy also degrades, especially for 9x9 Sudokus. The ablation 
    study of replacing the cGAN with a (weaker) standard  learnable decoder, without prior knowledge about single digits (DRNets w/o cGAN) shows that both the Sudoku and digit accuracy degrades dramatically.
  }  \label{table:mnist}
\end{figure}

\clearpage

\noindent
\clearpage

\newpage
\begin{figure}[ht]
  \centering
  \includegraphics [width=1\linewidth]
  {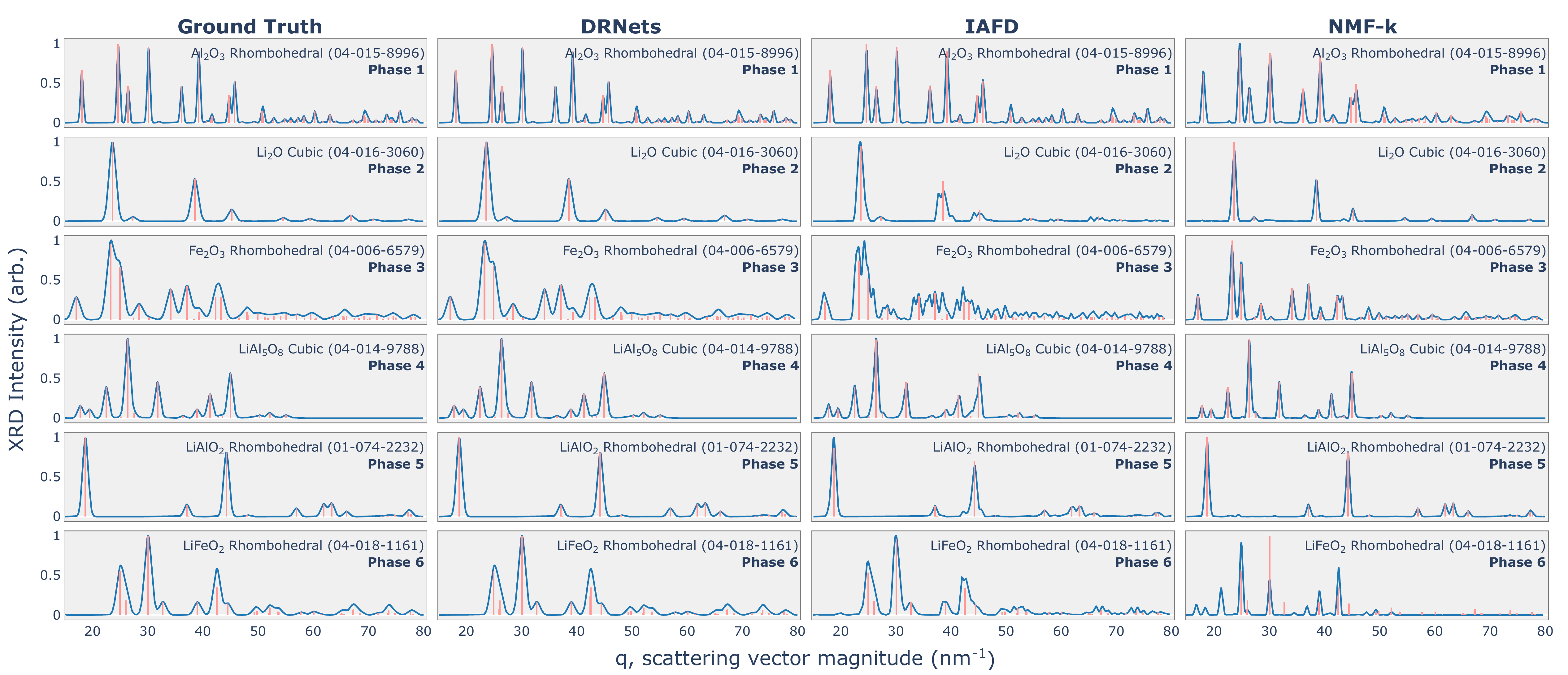} 
  %{FIGS/al-li-fe-phase-comparison3.png} 
  \caption{\textbf{Comparison of the phase patterns discovered by
      different methods vs.\ the ground truth phases for the Al-Li-Fe
      oxide system.} For each phase we plot the pattern of the
    recongized phase and the ICDD stick patterns.  While the
    phases discovered by DRNets closely match the ground truth phases,
    some of the IAFD and NMF-k's phases do not match well the ground
    truth phases (e.g., phase 3 (IAFD) and phase 6 (NMF-k)) as also
    reflected in the phase fidelity loss (0.00002 (DRNets);
    11.920 (IAFD); and 46.156 (NMF-k)); see also
    Fig.~\ref{fig:down-scale-PM}a). }
    \label{fig:Li-Fe-Al-phases}
\end{figure}
\noindent
\clearpage
\newpage

% \clearpage
% \noindent

\begin{comment}
\begin{figure}[ht]
%  \centering
\includegraphics[width=1\linewidth]{FIGS/Bi-Cu-V-phase-diagram_GMM.png} 
\caption{\textbf{Comparison of the activation map and the heatmap of L1 reconstruction loss for different methods for the Bi-Cu-V oxide system:} Each row denotes the activation of the different phases for the the different methods.
 Though we do not have  ground truth for the Bi-Cu-V oxide system, the solution generated by DRNets satisfies all thermodynamic rules with excellent reconstruction performance.
The heatmap on the right shows that DRNets reconstruct the XRD measurements much better than other methods with respect to the (log scale) L1 reconstruction under physical constraints of decomposed phases.
In addition, materials science experts thoroughly checked DRNets' solution of Bi-Cu-V oxide system, approved it, and subsequently discovered a new material that is important for solar fuels technology. See also Extended Data  Table \ref{table:PM_comparison}. }
    \label{fig:Bi-Cu-V-map}
\end{figure}
\noindent
\clearpage
\end{comment}

\newpage
\clearpage

\clearpage%HERE
\thispagestyle{empty} 
\begin{figure}[ht]
  \centering
  \includegraphics [width=1
    \linewidth]{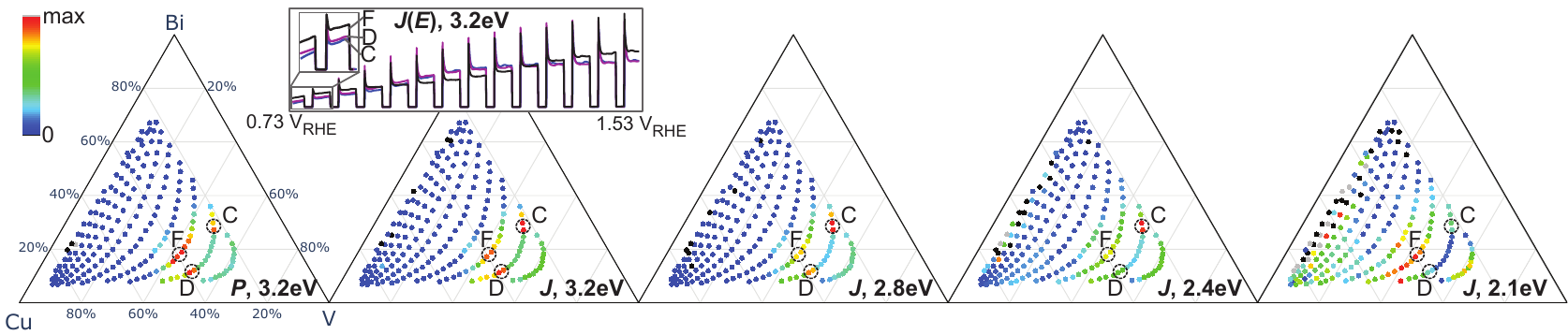} 
  \caption{\textbf{Characterization of Bi-Cu-V oxide library for photoelectrocatalysis of the oxygen evolution reaction, a critical reaction for solar fuels technology.}
  After XRD and XRF measurements, a grid of compositions was characterized with chronoamperometry (CA) with 4 different light emitting diode (LED) illumination sources from which photocurrent (\textit{J}) is calculated, as well as cyclic voltammetry with 3.2 eV illumination (CV) from the photoelectrochemical power generation (\textit{P}) is calculated. The resulting 5 performance metrics are plotted with respect to composition, and select pairs of points from 3 different phase fields in Extended Data Figure~\ref{fig:extra-bi-cu-v-o-drnet} are indicated with labels C, D and F. The common false color scale from 0 to a maximum value is used for each metric, with maximum values of 1.8 mW cm$^{-2}$ for \textit{P} and 13.3, 14.1, 0.5, 0.045 mA cm$^{-2}$ for \textit{J} with 3.2, 2.8, 2.4 and 2.1 eV illumination, respectively. The anodic sweep of the CV is shown for 3 select samples labeled by their phase region. All 3 of these regions contain BiVO$_4$, a well-known metal oxide photoanode, with much higher Cu concentration than typical Cu-free BiVO$_4$ photoelectrocatalysts. All 3 noted phase regions contain BiVO$_4$ and Cu$_3$(VO$_4$)$_2$ with D and F additionally containing Cu$_2$BiVO$_6$ and Cu$_2$V$_2$O$_7$, respectively. The different compositions and phase combinations lead to different performances, in particular the 3 phase region F exhibits higher photocurrent at low applied bias (see inset) and higher photocurrent with 2.1 eV illumination, which are 2 critical properties for BiVO$_4$ photoanodes that have been historically difficult to optimize. Despite common belief that phase mixtures are deleterious to photoactivity, these results demonstrate alloying and optimal phase mixtures as promising directions for photoanode discovery and optimization.
  %See details in \hyperref[sec:methods]{Methods}
   %
    \label{fig:bi-cu-v-pec}
    }
\end{figure}

\clearpage

\begin{figure}[ht]
\centering
  \includegraphics [width=1
    \linewidth]{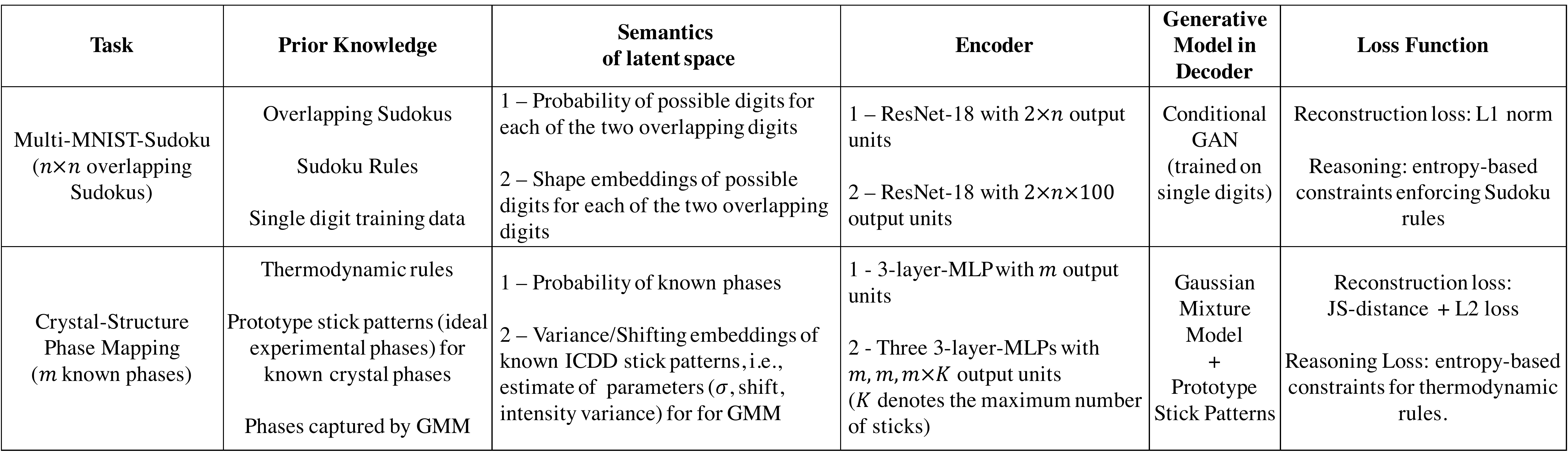} 
\caption{\small{\revised{\textbf{Different components of DRNets for the different tasks.}}}
      }
  \label{fig:table}
\end{figure}

\clearpage

%% file: si.tex
\section*{Supplementary \Carlanew{Note}}
\label{sec:note}
\noindent
\textbf{\Carlanew{Performance of DRNets for Multi-MNIST-Sudoku: }}
\Carlanew{DRNets significantly outperform CapsuleNet and ResNet, 
even when these supervised systems are coupled with local search to incorporate Sudoku rules.
This is particular true when going from 4-by-4 to  9-by-9 Multi-MNIST-Sudoku
instances:  DRNets' digit and Sudoku accuracy for 9-by-9 instances is close to  100\%, while ResNet's Sudoku accuracy is around 35\%. More comprehensive comparisons are provided in Extended Data Fig.~\ref{table:mnist}.
%and 80\% when boosted by local search. Also, while for the 4-by-4 we could improve the Sudoku accuracy of the supervised  methods by explicitly considering all the possible Sudoku solutions and selecting the most likely, such a strategy is out of the question for 9-by-9 instances, given that there are around $6.67\times10^{21}$ possible Sudoku solutions, which highlights the combinatorial nature of this problem.
}

\noindent
\textbf{\Carlanew{Ablation studies for Multi-MNIST-Sudoku: }}
We performed ablation studies of DRNets on the Multi-MNIST-Sudoku task to demonstrate the importance of the reasoning module and the generative decoder.
As shown in the Extended Data Fig.~\ref{table:mnist}, if we remove the reasoning modules, both the Sudoku accuracy and the digit accuracy would drop significantly, which shows that not only does the reasoning module help the Sudoku accuracy,  it  also improves the digit accuracy by eliminating impossible digits based on Sudoku rules. 
This effect is especially significant when it comes to the 9x9 case. 
In the 9x9 case, we used hand-written letters A-I for the second Sudoku in each overlapping Sudokus, which considerably increase the difficulty of recognizing the digit and the letter purely based on the reconstruction loss. 
Therefore, once we removed the reasoning module, the Sudoku accuracy dropped from 99.2\% to 0\% and the digit accuracy also dropped from 99.9\% to 63.8\%.
On the other hand, if we replace the generative decoder (cGAN) with a (weaker) standard learnable decoder, without prior knowledge about single digits, 
we can no longer connect the semantics of our latent space to the output of the decoder. 
Therefore, the optimization process can no longer find the right semantics for the latent space, even though the discovered nonsensical digits still follow the Sudoku rules (1.24\% digit accuracy, 0\% Sudoku accuracy).

\revised{
To demonstrate the importance of the data-driven learning by optimizing the model over multiple instances using shared parameters, we performed ablation studies of the "down-scalability" of DRNets, i.e.,  when optimized using only a few data instances (Extended Data Fig.~\ref{fig:down-scale-mnist}).
We evaluated the performance of DRNets with different dataset scales ranging from only one instance to 10,000 instances. The study shows the importance of learning the shared parameters across instances. Nevertheless,  DRNets  can reach 99\% accuracy with only 100 (unlabeled) 9x9 Multi-MNIST-Sudoku instances, a considerable smaller amount of data compared to standard deep learning approaches.
Since optimizing DRNets on a few instances may result in getting stuck at some local minimals such that the constraints are not all satisfied, we applied a restart mechanism \citeS{gomes1998boosting} on Multi-MNIST-Sudoku to circumvent this issue. 
Specifically, since DRNets directly incorporate logical constraints, we can check whether those constraints are satisfied or not at the end of a run. 
If not, for instances with violated constraints, we re-run the algorithm again on them.
In this ablation study, we applied at most 3 "restart" runs for each datasets to get the reported results on Extended Data Fig.~\ref{fig:down-scale-mnist}.
Due to the restart mechanism, the performance of DRNets on 10,000 instances are slightly better than the one without restart (Extended Data Fig.~\ref{table:mnist}).
Moreover, for the datasets with small scales (<1000), we performed multiple runs with randomly sampled instances to obtain an averaged performance.
}

\section*{Supplementary \Carlanew{Methods}}
\label{sec:si}

%\subsection{Applications}
%Instantiation
% {\color{blue} ChenDi, this application section needs work. As we
% discussed, it needs to be written in a way that there is: some deep
% learning lingo for archicteture stuff, data description, a bit of
% experimental confguration and training, but not too much given the
% space limitations. So we need to find a balance between wha goes here
% and what goes in the SI. I do like the summary you have in the main of
% the ICLR paper}
% {\color{blue} I think the rest of the blue text goes in the
% SI. All the experiments are performed on one NVIDIA Tesla V100 GPU
% with 16GB memory.  For the training process of our DRNets, we select a
% learning rate in $\{0.0001, 0.001\}$ with Adam
% optimizer\citeS{kingma2014adam}, for all the experiments. For baseline
% models, we followed their original configurations and further
% fine-tuned their hyper-parameters to saturate their performance on our
% tasks.}
\subsection{Multi-MNIST-Sudoku}
%\label{subsubsec:Sudoku}
\label{SI:Sudoku}

% {\color{blue} ChenDi  this section has to be re-written.}\\
%%%%%%%%%%%%%%%%%%%%%%%%%%%%%%%%%%%%%%%%%%%%%%%%%%%%%%%%%%%%%
\noindent
{\textbf{Data description:}}
For Multi-MNIST-Sudoku, we generated $n^2\times10,000$ (n is 4 or 9) input data points for each training set, validation set and test set, where each data point corresponds to a 32x32 image of overlapping digits/letters  from MNIST \cite{lecun1998gradient} and EMNIST~\cite{cohen1702emnist} and every batch of $n^2$ data points forms a n-by-n overlapping Sudokus.
For the 9x9 case, to distinguish the two overlapping Sudokus, we used letters A-I from EMNIST 
%as the digits 1-9 
for the second Sudoku. 
For ease of presentation, we may still refer to letters as "digits" in the following content.
We generated an extra \textit{cGAN} dataset, which is composed of 25,000 \textit{original MNIST/EMNIST images} for training the conditional GAN. Note that these four datasets are generated using disjoint sets of digit images.
%
% For Visual Sudoku, we generated $81\times10,000$ input data points, where each data point corresponds to a 32x32 hand-written digit image and every batch of $81$ data points forms a standard 9x9 Sudoku. We used digits 1 to 9  to denote  known cells and  digit 0 for the empty/missing cells.
% Similarly, we pre-trained a conditional GAN using digits 0 to 9 from a disjoint \textit{cGAN} dataset as the generative decoder.

\noindent
 %Di just changed the order - the encoder first
\textbf{DRNets for 
%Sudoku Games:}
Multi-MNIST-Sudoku:}
For Multi-MNIST-Sudoku, the encoder is made of two ResNet-18 models \cite{he2016deep} adapted from the PyTorch source code.  
The output layer for the first network has $2n$ (n is 4 or 9) dimensions, which models the two distributions $P_i$ and $Q_i$ for the two overlapping digits.  
Another network outputs $2n$ 100-dimensional ($200n$ dimensions in total) latent encoding $\mathbf{z}_{i,j}$ to encode the shape of the possible $2n$ digits conditioned on the input mixture, and is used by the generative decoder to generate the reconstructed digits.
%$G(z_{i,j})$.
\revised{
We use a conditional GAN \cite{mirza2014conditional}, 
as the generative model in our generative decoder (for ease of presentation, we abuse a bit the notation of $G(\cdot)$ to also denote the generative model)}, which is adopted from the implementation of \citeS{eriklindernoren2019} and pre-trained using digits in the \textit{cGAN} dataset.
Note that this is the only supervision we have in this task, which is even weaker than the general concept of the \textit{weakly-supervised setting} \citeS{zhang2017weakly}.
\revised{
Given the \textit{shape} ($\mathbf{z}_{i,j}$) and \textit{probability} embeddings ($P_i$ and $Q_i$), DRNets 
estimate the two digits in the cell by computing the expected digits over $P_i$ and $Q_i$, i.e., $\sum^n_{j=1}{P_{i,j}G(\mathbf{z}_{i,j})}$ and  $\sum^n_{j=1}{Q_{i,j}G(\mathbf{z}_{i,j+4})}$, and remix them to reconstruct the original input mixture (Fig.~\ref{fig:sudoku}g-i and % Extended Data
Fig.~\ref{fig:latent_space}a).}
We use entropy-based functions to impose the continuous relaxation of the cardinality and All-Different constraints to reason about the  Sudoku structure, \Carla{which results in a total of  $2n^2+6n$ constraints for the nxn Sudoku ($n^2\times2$  cardinality constraints enforcing a single digit per cell; $n\times3\times2$  all-different constraints enforcing no repetition of a digit in a row, column, and box 
%among cells of the overlapping Sudokus
(see Extended Data Fig.~\ref{fig:DRNet-flow}b)}.

\noindent
%\subsection{Continuous Relaxations for Multi-MNIST-Sudoku:}
\textbf{Continuous Relaxations for Multi-MNIST-Sudoku:}
%\label{SI:Relaxations-Sudoku}
In Multi-MNIST-Sudoku, there are two types of constraints in the Sudoku rules: (1) \textit{cardinality constraints} and (2) \textit{All-different constraints}.

\noindent
\textit{\textbf{Cardinality Constraints:}} One straightforward encoding of the discrete version of cardinality constraints, which are used to encode the possible two digits in the cell indicated by data point $x_i$ (one from $\{1...4\}$ and the other from $\{5...8\}$), is to use 8 binary variables $e_{i,j}\in\{0, 1\}$, while requiring $\sum^4_{j=1}e_{i,j} = 1$ and $\sum^8_{j=5}e_{i,j} = 1$.
To relax the discrete variables, DRNets model probability distributions $P_i$ and $Q_i$ over digits 1 to 4 and 5 to 8 respectively:  $P_{i,j} ,\scriptstyle{j=1...4}$ and $Q_{i,j}, \scriptstyle{j=1...4}$  denote the probability of digit $j$ and the probability of digit $j+4$, respectively.
Then, we can approximate the cardinality constraint of $e_{i,j}$ by minimizing the entropy of $P_i$ and $Q_i$, which encourages $P_i$ and $Q_i$ to collapse to one value.
One can see, when the entropy of distributions $P_i$ and $Q_i$ reaches 0, all the probability mass collapses to only one variable. 
Therefore, all $P_{i,j}$ and $Q_{i,j}$ are either 0 or 1, which is a valid solution of the original discrete constraints. 
\begin{comment}
\begin{figure}[h]
    \centering
    \includegraphics[width=1\linewidth]{FIGS/relaxation.pdf}
    \caption{ 
    % Examples of continuous relaxations: $e_{i,j}, N_c,
    %   N_l,K_j,\lambda_h$, $P_i$ denote binary variables, the number of
    %   clauses, the number of literals, the number of literals in the
    %   $j$-th clause, the weights of entropy terms, and the Bernoulli
    %   distribution for the $i$-th literal.  "leaky\_relu" is the leaky
    %   ReLU.%, where we let $x_i$ be the probability that $x_i$ is true
    Examples of continuous relaxations: $e_{i,j}, P_i, Q_i, P_M, N_c, N_l,K_j,\lambda_h$, $B_i$ denote binary variables, the discrete distribution over digits 1 to 4, the discrete distribution over digits 5 to 8, the discrete distribution over values 1 to $M$, the number of clauses, the number of literals, the number of literals in the $j$-th clause, the weights of entropy terms, and the Bernoulli distribution for the $i$-th literal.  
    "leaky\_relu" is the leaky ReLU and let $\bar{x}_i$ be the probability that $x_i$ is false.  }
    \label{fig:SI_relaxation}
\end{figure}
\end{comment}

\noindent
\textit{\textbf{All-different Constraints:}}
Another combinatorial constraint %that we face 
in Multi-MNIST-Sudoku is the All-Different constraint, where all the cells in a \textit{constrained set} $S$, i.e., each row, column, and any of four 2x2 boxes involving the corner cells, must be filled with non-repeating digits.
For a probabilistic relaxation of the All-Different constraint, we analogously define the entropy of the averaged digit distribution for all cells in a constrained set $S$, i.e., $H(\bar{P}_S):$
%\begin{linenomath}
\begin{equation}
    %\resizebox{0.7\textwidth}{!}{$
    H(\bar{P}_S) =-\sum^4_{j=1}\bar{P}_{S,j}\log{\bar{P}_{S,j}}=
    -\sum^4_{j=1}\bigg(\frac{1}{|S|}\sum_{i\in S}P_{i,j}\bigg)\log\bigg({\frac{1}{|S|}\sum_{i\in S}P_{i,j}}\bigg)
    %$}
\end{equation}
%\end{linenomath}
In this equation, a larger value implies that the digits in the cells of $S$ distribute more uniformly.
Thus, we can analogously approximate All-Different constraints by maximizing $H(\bar{P}_S)$ and $H(\bar{Q}_S)$ while minimizing all $H(P_i)$ and $H(Q_i)$, $i\in S$ (for the cardinality constraints of cell-$i$).
When the entropy of the digit distribution in each cell is 0, we know that the digit distribution of each cell converges to one digit. 
Hence, if $H(\bar{P}_S)$ reaches its maximum, i.e., $\log|S|$, we have $\frac{1}{|S|}\sum_{i\in S}P_{i,j}=\frac{1}{|S|}$ for all digit $j$. 
Crossed with the fact that $P_{i,j}$ are either 0 or 1 when the cardinality constraints are satisfied, we know that only one $P_{i,j}$ is equal to 1 for all cell $i$ in the set $S$ and others are 0, which directly states the All-Different constraints.

\noindent
\textbf{Baselines for Multi-MNIST-Sudoku:}
%See Extended Data Table \ref{table:mnist} for a summary of results and  \hyperref[SI:Sudoku]{SI~\ref{SI:Sudoku}}.
We compared DRNets with
the state-of-the-art for demixing digits, which are supervised methods: CapsuleNet \cite{sabour2017dynamic} and ResNet \cite{he2016deep}. Though ResNet and CapsuleNet have access to labeled data, they do not utilize Sudoku rules. 
Therefore, to saturate their performance, we further imposed Sudoku rules into those baselines via a post-process.
Specifically, for 4x4 cases, we did a local search for the top-2 (top-3 would take too long to search) most likely choice of digits for each Sudoku of the two overlapping Sudokus and try to satisfy Sudoku rules with minimal modification compared with the original prediction.
Since there are only 288 different 4x4 Sudokus, we also performed an exhaustive search, which explicitly considers all the possible Sudoku configurations and selects the most likely one. 
However, such a strategy is out of the question for 9-by-9 instances, given that there are around $6.67\times10^{21}$ possible 9x9 Sudokus.
For 9x9 cases, we conduct a different local search algorithm since enumerating the top-2 digits for an 9x9 Sudoku is not feasible ($2^{81}$). 
We conducted a depth-first search for digits in each cell from most likely to less likely until it finds a valid Sudoku.

Because CapsuleNet \cite{sabour2017dynamic} did not provide a source code, we adopted the implementation of Laodar \citeS{laodar2017}, with minor modifications. 
For ResNet, we adopted the ResNet-18 architecture \cite{he2016deep} and trained it in a multi-class classification setting, where we provide explicit supervision for the labels of two digits (one from \{1..4\} and the other from \{5..8\}) in each cell. 
We did a grid search for the hyper-parameters of both baseline models to saturate their performance. 
Extended Data Fig.~\ref{table:mnist} shows the comparison of the performance of different methods for Multi-MNIST-Sudoku %(Test set: 10,000 instances)
showing how DRNets outperforms other approaches and how it even has the capbility of self-learning by reasoning about the Sudoku rules.

For the training of DRNets, we used $L1$ loss as the reconstruction loss between the reconstructed mixture and the original input. 
For the initial weights of constraints, we set 0.01 for the cardinality constraints, 1.0 for the All-Different constraints, and 0.001 for the $L1$ loss. 
Note that, DRNets are really "self-supervised" \citeS{jing2019self} by the Sudoku rules and the self-reconstruction, instead of the standard supervision by labeled data.
Therefore, we can directly use DRNets to solve the test set without using a training set.
We optimize DRNets on the test set for 100 epochs with a batch size of 100, and it took 50 minutes to finish and achieve the reported performance for the 10,000 overlapping Sudokus.

\subsection{Crystal-Structure Phase-Mapping}
\label{Method:phase_mapping}%\label{SI:phase_mapping}
We illustrate the DRNets for crystal structure phase mapping for two chemical systems: (1) a ternary \textbf{Al-Li-Fe}
oxide system, described in Le Bras et.\ al.~\cite{le2014challenges}, which is 
synthetically generated from a known phase diagram, which also provides the  ground-truth solution, and (2) an experimental system from a continuous composition spread thin film from the ternary \textbf{Bi-Cu-V} oxide system.

For each system, the input data points are XRD patterns, each containing signals from 1 to 3 phases, as well as the composition of each system. Each XRD pattern is the (nonnegative) XRD scattering intensity as a function of the scattering vector magnitude, $Q$. A peak in the XRD pattern results from Bragg scattering and indicates the presence of a plane of atoms in the crystal structure with interplanar spacing of $d=2\pi /Q$. Alloying refers to chemical substitution within a crystal structure where a change in composition causes the crystal structure to stretch or contract, so an expansion by 1\% would cause the $d$ values of the peaks to ``shift'' multiplicatively by 1\%, corresponding to 1\% decrease in the $Q$ value of each peak. Alloying can occur with negligible expansion/contraction of the crystal structure, in which case it cannot be directly detected by measured peak positions.  Complex alloying may occur where alterations to the aspect ratio of the crystal structure causes nonuniform peak shifting, which can be modelled by DRNets but was not in the present work. A phase diagram comprises a graph of the regions of composition space where each combination of phases is observed (the phase fields), as well as annotation of any alloying that occurs within each region.\citeS{smith_chapter_2007} DRNets provide all phase fields within the hull of input XRD pattern compositions, as well as peak shifting for each phase that captures most instances of alloying.

Ternary compositions are plotted in a standard 2-D Euclidean triangle plot and Delaunay triangulation provides edges representing neighboring composition points, which is used to establish the mathematical expression of the thermodynamic rules. 
The mathematical descriptions of our implementations of these rules are
\Carlanew{described below. The rules }
%provided in the 
%\hyperref[SI:phase_mapping_relaxations]{SI~\ref{SI:phase_mapping_relaxations}}, 
%and here we note 
%briefly describe the physical meaning of the constraints. The 
%that the rules all 
result from considering free energy thermodynamics in the context of a composition space with pressure and temperature held constant, as they were in the synthesis of the materials. The names and brief description of the rules are as follows: ``Gibbs'' is based on the standards Gibbs' Phase Rule and limits the number of phases that can coexist; ``Gibbs-Alloy'' is an extension of the same thermodynamic rule where the thermodynamic degree of freedom assumed by alloying lowers the max phase count; ``Phase-Field-Connectivity'' is the composition graph implementation of the definition of a phase field in a phase diagram, i.e. that each phase field comprises a continuous region of composition space. These thermodynamic rules are central to phase diagram determination\citeS{smith_chapter_2007,edwards_chapter_2007} and have been implemented in different ways in our previous work,\cite{gomes2019crystal}  
%and in part by [Kusne et al. 2015]\citeS{kusne2015high}.
and incorporated to some extent in Refs \cite{stanev2018unsupervised} and \cite{kusne2015high} (e.g. with sparsity constraints and clustering that lowers the number of  activated phases).

For the Bi-Cu-V oxide
system, the thin film materials were prepared by sputter co-deposition from Bi, Cu, and V sources. The positions of the sources form an equilateral triangle with a substrate above the center on the triangle collecting atoms at different rates at each position, as described previously \citeS{suram2015combinatorial}. 
At each substrate location the unique mixture or composition of Bi, Cu and V atoms are mixed at the atomic level and crystallize into crystal domains generally between 5 and 100 nm upon subsequent annealing (550 °C in air in this case), forming a thin film approximately 200 nm in thickness.
For a given 1 mm$^2$ area representing a given composition, the mixture of order 10$^{10}$ crystal domains comprising the
%are  typically of 1 to 3 different phases, which are characterized through synchrotron x-ray diffraction\citeS{gregoire2014high} to generate the XRD pattern for each composition.
1 to 3 different phases are characterized through synchrotron x-ray diffraction\citeS{gregoire2014high} to generate the XRD pattern for that composition.

\revised{There are 307 composition data points for Bi-Cu-V system and each XRD pattern contains $D=300$ diffraction signals equally spaced from $Q$=5 to 45 nm$^{-1}$.
There are 231 composition data points for the benchmark Al-Li-Fe system and each XRD pattern contains $D=650$ \revised{diffraction signals} equally spaced from $Q$=15 to 80 nm$^{-1}$. Each XRD pattern is normalized to a maximum intensity of 1. 
The set of prototype stick patterns for each system correspond to known crystal structures from the International Centre for Diffraction Data (ICDD) database. %The database entry numbers are provided in the SI.
}

%%%%%%%%%
\noindent
\textbf{DRNets for Phase-Mapping:}   
%ChenDi - succint description of the encoding neural network 
Given the input $D$-dimensional XRD pattern,
%vector representing the intensity of the mixture of XRDs at different diffraction angles,
we use four 3-layer-fully-connected networks as our encoder to encode a two-part latent space, which captures the probabilities (denoted as $P_{i,j}$) and shapes of the possible phases (denoted as $\mathbf{z}_{i,j}$) and is constrained by the reasoning module to satisfy the thermodynamic rules 
%(Extended Data Fig.~\ref{fig:DRNet-latent}b).
(Fig.~\ref{fig:latent_space}).
To model more realistic conditions, the generative decoder of the DRNets uses Gaussian mixture models \citeS{lindsay1995mixture} to approximate the  
measured phase patterns %from \textit{stick patterns}
where the relative peak locations and 
intensities
%mixture coefficients 
are given by the prototype stick pattern, and 
the latent encoding $\mathbf{z}_{i,j}$ parameterizes the other information needed to simulate an XRD pattern with a series of Gaussian peaks: the peak width, multiplicative shift of peak locations, and possible amplitude variance. The set of phase activations sum to 1 and are the relative intensities of the set of generated phase-pure patterns whose sum is the reconstruction of the input XRD pattern. To remove negligible activations that are mainly caused by experimental noise we
applied a simple post-processing that cuts-off all the activations that are lower than 1.0\%.
The
%by the stick locations and amplitudes and the peak width, multiplicative location shift, and possible amplitude variance are parameterized by the latent encoding $z_{i,j}$.
%The overall objective function of the DRNets combines responses from the generative decoder   and the reasoning module, which is optimized using constraint-aware stochastic gradient descent.
%Specifically, 
output of the first three networks in the encoder are $M$-dimensional vectors: $(P_{i,1}, ..., P_{i,M})$, 
$(\alpha_{i,1}, ..., \alpha_{i,M})$, 
$(\sigma_{i,1}, ..., \sigma_{i,M})$
($M$ is the number of possible phases, e.g., 159 for the Al-Li-Fe oxide system), which represent the probability $P_{i,j}$ of each phase-$j$ at data point $i$, the multiplicative shifting ratio $\alpha_{i,j}$, and the standard deviation $\sigma_{i,j}$ of the Gaussians characterizing the peaks in phase-$j$, respectively.
The output of the last network is a $M\times K$-dimensional vector, representing the possible amplitude variance of peaks in each phase. Here, $K$ is the number of maximal peaks in a stick pattern ($K=200$).
For the first 3 networks there are 1024, 1024, and 512 hidden units, respectively per layer, and  for the last network there are 512, 512, and 32 hidden units, respectively per layer. The last layer has fewer hidden units given its high-dimensional output space ($M \times K$).
All networks use ReLU~\citeS{nair2010rectified} as their activation function.

% The structured latent encoding is used by a generative decoder, a Gaussian mixture models [ref], to
% reconstruct the phases. 

%Imposing thermodynamic rules is challenging, especially when constraints, such as \textit{phase field connectivity} and \textit{Gibbs-alloying rule}, potentially concern all data points in
%the composition graph. 
The thermodynamic rules can concern many to all points in the composition graph, creating a challenging set of combinatorial constraints to impose.
In Multi-MNIST-Sudoku, where each overlapping Sudoku naturally forms the maximal connected components in the \textit{constraint graph},
%(see definition in \hyperref[SI:SGD]{Methods})
we can easily batch every $n^2$ data points together to reason about the All-Different constraints among them. However, in Crystal-Structure-Phase-Mapping, since the maximal connected component involves all data points in the composition graph, neither batching all data points into the memory nor reasoning about the whole graph is tractable.  
Therefore, to enforce the connectivity constraint we devised a strategy of sampling the large connected component through many local structures (still connected components) and solve each of them iteratively.  
Specifically, for each oxide system,
we sampled 100,000 paths in the composition graph via Breadth First
Search to construct a path pool.  Then,  for every iteration, DRNets
randomly sample a path from the pool and batch the data points
along that path (see Fig.~\ref{fig:bi-cu-v-o}b).  Finally, we only reason
about the thermodynamic rules along the path.  By iteratively solving
sampled local structures (paths) of the "large" maximal component, we
can cost-efficiently approximate all global constraints. 
%We provide additional details concerning the implementation of the thermodynamic rules in  \hyperref[SI:phase_mapping_relaxations]{SI~\ref{SI:phase_mapping_relaxations}}.  

\begin{comment}
\noindent
\textbf{Training of DRNets for Phase Mapping.} 
% Given that we only have limited amount of data points in each oxide system, we only consider the optimization mode of DRNets, 
% Therefore, 
We directly optimize DRNets' performance on each system via constraint-aware stochastic gradient descent.
%(see Fig.~\ref{fig:bi-cu-v-o} and Extended Data Fig~\ref{fig:DRNet-flow}).
See detailed experimental information and algorithmic details in \hyperref[SI:phase_mapping]{SI~\ref{SI:phase_mapping}}.
\noindent
\textbf{Baselines:}  
See 
%Fig.~\ref{fig:Li-Fe-Al-map},
Fig.~\ref{fig:phase-diagram-main}b,
Extended Data Fig.~\ref{fig:Li-Fe-Al-phases}, and
Fig.~\ref{fig:down-scale-PM}a for a summary of results for the Al-Li-Fe
oxide system and
Fig.~\ref{fig:phase-diagram-main}a, Fig.~\ref{fig:extra-bi-cu-v-o-drnet}, and
Fig.~\ref{fig:phase-diagram-main}a
for the Bi-Cu-V oxide system and \hyperref[SI:phase_mapping]{SI~\ref{SI:phase_mapping}}.
\end{comment}

%\subsection{Continuous Relaxations for Crystal-Structure Phase Mapping:}
%\label{SI:phase_mapping_relaxations}
%In Crystal-Structure Phase Mapping, we imposed the following thermodynamic rules in DRNets:
\noindent
\textbf{Continuous Relaxations for Crystal-Structure Phase Mapping:}
The thermodynamic rules imposed in DRNets are based on standard properties of isothermal compositional diagrams\citeS{smith_chapter_2007,edwards_chapter_2007}. The only constraint applied to each XRD pattern independently is the maximum of 3 phases. This is a consequence of Gibbs' phase rule where the 2 composition degrees of freedom are the thermodynamic variables corresponding to a maximum phase count of 3 at any point within the ternary composition space. An extension of this rule is that alloying removes at least 1 thermodynamic degree of freedom and thus lowers the maximum phase count to 2. Alloying is detected through comparison of the multiplicative shifting parameters (in  the DRNets latent space), between neighboring compositions, where a shifted pattern between 2 neighbors that share the same set of phases invokes the Gibbs-Alloy rule where both patterns may contain only 2 phases. The final rule relates to the compositional connectivity of each phase fields, i.e. Phase-Field-Connectivity. Thermodynamic compositional phase diagrams comprise the convex hull of the Gibbs free energy for all phases as a function of composition. The generally parabolic shape of the free energy for each phase implies that if a phase or set of phases appears on the convex hull at one composition, every other composition where it appears can be accessed by a contiguous composition path. Further the amount of the phase, i.e. its activation in the XRD patterns, will vary smoothly in composition space. Importantly, the thin film synthesis of the Bi-Cu-V system can result in non-equilibrium phase behavior. In our experience, from inspection of thousands of XRD patterns of thin films, any deviation from equilibrium typically does not alter the rules as enforced, likely because the alterations correspond to a kinetically impeded phase that is removed from ``consideration'' in the free energy diagram, or alteration of the effective free energy for a phase, both of which  can change which phases are experimentally observed but not the enforced rules. Indeed this is a key reason why phase mapping of experimental systems is required, as opposed to relying on computed free energy diagrams. 

\begin{comment}

\noindent
We also highlight that, as we noted in the methods, DRNets uses a batching strategy that samples,  for each oxide system,
 100,000 paths in the composition graph via Breadth First
Search to construct a path pool.  Then,  for every iteration, DRNets
randomly sample a path from the pool and batch the data points
along that path (see Fig.~\ref{fig:bi-cu-v-o}). DRNets, reason
about the thermodynamic rules only along the randomly sampled path pool.
\end{comment}

\noindent
\textbf{Implementation of thermodynamic rules for Crystal-Structure Phase Mapping:}
\noindent
\textit{\textbf{Gibbs:}} This rule states the maximum number of co-existing phases, which is imposed via the relaxation of the k-sparsity constraints. 
For the discrete version, we can model the existence of $M$ possible phases of $i$-th data point using binary variables $e_{i,j}$ ($j=1...M$), while requiring $\sum^M_{j=1}e_{i,j}\leq k$.
We derive the relaxation of k-sparsity constraints in a similar way as the cardinality constraints except that we now want to force the distribution to concentrate on at most k entities (phases).
By normalizing the values of discrete variables $e_{i,j}$ ($j=1...M$) to a discrete distribution $P_M$, we can minimize the entropy of distribution $P_M$ to at most $\log k$, which is the maximal entropy when the distribution concentrates on only $k$ values. 
Though, $H(P_M)<\log k$ is not a sufficient condition for k-sparsity, we can initialize the threshold $c$ of k-sparsity constraints to $\log k$ and dynamically adjust the value of $c$ based on the satisfaction of the k-sparsity constraints. 
In practice, it works well with the supervision from other modules, such as the self-reconstruction.

\noindent
\textit{\textbf{Gibbs-Alloy:}} 
%This rule states that if "alloying" happens, then the maximum number of possible co-existing phases should decrease by one. "Alloying" is a phenomenon that the stick locations of a phase (crystal structure) shift (change) with respect to composition neighbors.
DRNets explicitly model the shifting ratio in the generative decoder and penalize the difference between adjacent data points along our sampled path.  The reasoning module keeps track of the difference of shifting ratio between adjacent data points, and when it is larger than a threshold (0.001), we confirm the existence of "alloying" and reduce the maximum number of possible co-existing phases by one via adjusting the threshold $c$ in the k-Sparsity Constraints.

\noindent
\textit{\textbf{Phase-Field-Connectivity:}} 
%This states that the distribution (also referred as activation) of a phase field should form a connected component in the composition graph, and the variation of the activation of each phase should also be smooth (see Extended Data  Fig.\ref{fig:Li-Fe-Al-map}). (Herein, the phase field refers to the co-existence of a combination of phases, including the existence of a pure phase.) 
We impose this rule by penalizing the difference between the phase activation of adjacent data points $P_u$ and $P_v$ along the sampled path. 
In our implementation, we used L2-norm to penalize the difference.

%\section{Baseline Models and Additional Experimental  Information}

%\subsection{Multi-MNIST-Sudoku}
%\label{SI:Sudoku}

%\subsection{Crystal-Structure Phase Mapping}
%\label{SI:phase_mapping}
% \Di{
% \textbf{(i) Detailed Network structure:}

% }

\noindent
\textbf{Evaluation Criteria for Phase Mapping:}
\label{sec:phasemappingcriteria}
Our evaluation criteria (see Fig.~\ref{fig:down-scale-PM}a) include reconstruction losses, phase fidelity loss and the satisfaction of thermodynamic rules.
Note that, before evaluating the reconstruction losses, we fit the demixed patterns (for all methods) to the closest prototype using a model described previously\cite{le2014challenges} to exclude noise. 
We quantified the phase fidelity loss by  measuring the Jensen-Shannon distance (JS distance) between the demixed XRD pattern and the closest pattern simulated from a prototype. 
The motivation for using the JS distance metric for fidelity of demixed patterns is that the set of peaks and their locations are the most important characteristics of a phase pattern.
%Thus, we normalized the $D$-dimensional XRD pattern vectors into probability distributions over diffraction angles and measure the mismatch of "peaks" between them via the JS distance between their corresponding distributions.
We normalize the area under each XRD pattern to be 1 and use the JS distance metric (with $\epsilon$ of 1e-9 to avoid division by zero) to quantify the difference between two patterns, which has a large loss when peaks appear in one pattern but not the other.

\noindent
\textbf{Manual Solution Evaluation for Phase Mapping:}
The detailed explanation of the main text statement ``the presence of each phase was manually verified by matching each prototype peak to a signal in a corresponding XRD pattern and by confirming that measured XRD peaks in the pattern are explained by the DRNets solution'' is as follows: for powder XRD patterns, all peaks (with intensity above the detectability limit) in the prototype must be in the measured pattern, and all peaks in the XRD pattern must be explained by prototypes. Failure of the former indicates that one is considering a prototype of the wrong crystal structure, and failure of the latter could be due to the wrong prototype or an undetected phase, so this analysis can only be done once all phases are identified. This is precisely why complex phase mapping in high dimensional composition spaces is intractable for humans because the pure-phase XRD patterns are not known and the existence of peaks from multiple phases creates many different possible choice of prototype(s) that explain some to all of the measured peaks, especially under consideration of possible peak shifting of the prototypes. In many cases, a single XRD pattern can only be definitively solved by reasoning about XRD patterns of related compositions, so finding a logically consistent solution is very difficult.

Phase mapping problems are particularly challenging when the XRD data does not contain examples of pure phase  patterns, as is often the case in combinatorial materials research. In the Bi-Cu-V DRNets solution, only 2 of the 13 phases appear with activations above 80\%, demonstrating the ability of DRNets to identify prototypes that only appear in combination with other prototypes.
%max activations are 0.32,0.60,0.74,0.97,0.11,0.76,0.88,0.28,0.73,0.21,0.46,0.62,0.15

\Carlanew{In phase mapping, we also note the possibility of the presence of a phase in the input XRD patterns that is entirely distinct (not just modified) from all prototypes. We don't discuss handling such cases since it is beyond the scope of this work, requiring density function theory analysis, 
% because we don't have experimental demonstrations of them, 
although our results on reconstruction loss (DRNets produces much better data reconstruction than other methods) demonstrate that the presence of a new phase would be readily identifiable by an uncharacteristically high reconstruction loss in a specific composition region. }

\noindent
\textbf{Baselines for Crystal-Structure Phase Mapping:}
%\Di{\textbf{(ii) Baselines:}}
We compared DRNets with the state-of-the-art phase mapping algorithms, IAFD 
\cite{gomes2019crystal} and NMF-k \cite{stanev2018unsupervised}, which are both non-negative matrix factorization (NMF) based unsupervised demixing models. 
NMF-k improves the pure NMF algorithm \citeS{long2009rapid} by clustering common phase patterns from thousands of runs.
However, NMF-k does not enforce thermodynamic rules and therefore the solutions produced are often not completely physically meaningful.
%result generated by NMF-k often violates thermodynamic rules and cannot be considered as a valid solution. 
IAFD uses external mixed-integer programming modules to enforce thermodynamic rules during the demixing. 
However, due to the gap between the external optimizer and NMF module, the solution of IAFD is still far from the ground truth. 
%Extended Data Table~\ref{table:PM_comparison} shows the experimental results.

Our evaluation criteria (see Fig.~\ref{fig:down-scale-PM}a) include reconstruction losses, phase fidelity loss and the satisfaction of thermodynamic rules.
Note that, before evaluating the reconstruction losses, we fit the demixed phases (for all methods) to the closest ideal phases using the physical model \cite{le2014challenges} to exclude noise. 
Meanwhile, we quantified the phase fidelity loss by  measuring the Jensen-Shannon distance (JS distance) between the demixed phases and the closest ideal phases. 
The reason of using the JS distance to measure the fidelity is that the location of peaks are the most important characteristics of a phase pattern. 
%Thus, we normalized the $D$-dimensional XRD pattern vectors into probability distributions over diffraction angles and measure the mismatch of "peaks" between them via the JS distance between their corresponding distributions.
We normalize the area under each XRD pattern to be 1 and use the JS distance metric (with $\epsilon$ of 1e-9 to avoid division by zero) to quantify the difference between two patterns, which has a large loss when peaks appear in one pattern but not the other.

We also considered a recently proposed supervised algorithm~\cite{lee2020deep} in which a deep neural network, trained using simulated data based on known crystal phase patterns, directly predicts the phases present in a given XRD pattern. 
While such an approach can be effective for complementing human expertise for a single XRD pattern, it performed poorly on a complex system such as the benchmark Al-Li-Fe oxide system (phase identification accuracy around 1\%), which highlights the limitations of a purely simulated-based supervised approach for handling the combinatorics of phase mapping. 
Moreover, we note  that the method can only provide a classification for the existence or the quantized fractions of phases instead of 
%the true decomposed XRD patterns
actually estimating the weights (regression task) of the  phase activations,  as done by  DRNets.
%, we cannot directly compare its performance with other methods. 
So, we evaluated DRNets using the setting of their classification tasks on their Li-Sr-Al powder system  dataset.
We used a similar setting of hyper-parameters as for the Bi-Cu-V oxide system. 
We note that this powder system only contains one phase combination (Li$_2$O, SrO, and Al$_2$O$_3$): both their model and  DRNets achieved 100\% phase recognition accuracy. 
For the quantized phase fraction (0-33\%, 33\%-66\%, and 66\%-100\%) classification tasks, unsupervised DRNets have an accuracy of 95.3\%, outperforming their supervised model with accuracy of 86.0\%.
Beyond this quantized phase activation classification, DRNets can further provide estimates of the weights of the phase activations with an average error of 4.7\%.

%Note that, to avoid division by zero, we bounded the divisor by an $\epsilon=\mbox{1e-9}$.

%\Di{\textbf{(iii) Training Details:}}

In the optimization process of DRNets, we used the JS distance with a weight of 20.0 plus the L2-distance with a weight of 0.05 as the reconstruction loss.
Due to the different noise level in the Al-Li-Fe oxide system and the Bi-Cu-V oxide system, we use different weights for the penalty functions of different constraints.
For Al-Li-Fe oxide system, the weights of k-sparsity constraints (Gibbs rule) is 1.0 and the weights of phase field connectivity constraints is 0.01.
For Bi-Cu-V oxide system, the weights of k-sparsity constraints (Gibbs rule) is 30.0 and the weights of phase field connectivity constraints is 3.0.
%We used different weights (1.0 and 30.0) of k-sparsity constraints for Li-Fe-Al oxide system and Bi-Cu-V oxide system due to their different noise level.
%We used the L2-distance for phase field connectivity constraint with a weight of 
%(See our code for other hyper-parameters).
In terms of the optimization process, DRNets took about 30 minutes to achieve the reported performance for both systems. 
%\Di{In contrast, IAFD and NMF-k not only have a much worse performance w.r.t.\ the solution quality but also 
%take several hours to generate their solution. 
%}
In contrast, not only do IAFD and NMF-k have a much worse performance with respect to the solution quality, they   also 
take considerable longer, several hours, to generate their solutions. 
%IAFD and NMF-k have a similar time performance but a much worse performance w.r.t. the solution quality.
In fact, for the Bi-Cu-V oxide system, both NMF-k's solution and IAFD's solution are not physically meaningful.  

\noindent
\textbf{Down-Scaling Analysis for Crystal-Structure Phase Mapping:}
\revised{
We also investigated the "down-scalability" of DRNets for crystal structure phase mapping tasks, which further confirmed the importance of learning DRNets across multiple samples.
In this experiment, we used \textit{phase activation accuracy} as the evalution metric, which evaluates the percentage of sample points that have the same set of phases as the ground truth. 
For Al-Li-Fe oxide system, we evaluated the performance of DRNets learned over different number of XRD patterns ranging from one XRD patterns to the entire set. 
For the case of using K XRD patterns, we performed $\floor{231/K}$ runs of DRNets on the XRD patterns randomly sampled from the composition space to obtain an averaged performance. 
As shown in  Fig.~\ref{fig:down-scale-PM}, learning over multiple XRD patterns plays an important role for DRNets to solve the Al-Li-Fe oxide system and DRNets can almost perfectly recover the phase activation of XRD patterns when it is learned over more than 150 XRD patterns.
Since we do not have the ground truth of the Bi-Cu-V oxide system, we only evaluate the consistency between the results of running DRNets on all 307 XRD samples and the results of running DRNets on each XRD sample one at a time (named DRNet-single).
We made minor edition to the DRNet to adapt to the setting of running on a single XRD pattern.
For DRNet-single, we ignored the Gibbs-alloying rule and the phase-field-connectivity rule since they do not make sense on a single sample point.
As a result, DRNet-single can only identify the same phases as DRNets for 27\% of the patterns, highlighting that the nuanced phase behavior of this system can only be resolved through combinatorial experimentation matched with learning the shared parameters across multiple XRD patterns combined with reasoning about the underlying complex thermodynamic constraints.
}